%% file: example_paper.tex
\documentclass[aps,prd,reprint,groupedaddress,nofootinbib,]{revtex4-1}
\usepackage[dvipsnames]{xcolor}
\usepackage{graphicx}
\usepackage{hyperref}
\usepackage{url} 
\usepackage{ulem}
\usepackage{makecell}
\hypersetup{
    pdfnewwindow=true,      
    colorlinks=true,        
    linkcolor=blue,         
    citecolor=blue,        
    filecolor=blue,         
    urlcolor=blue           
}

\newcommand{\aap}{Astron. Astrophys.}

\newcommand{\apjl}{ApJ Lett.}
\newcommand{\apjs}{ApJS}

\usepackage{pifont}

\usepackage{amsmath}
\usepackage{amssymb}

\usepackage[acronym]{glossaries}
\setacronymstyle{long-short}
\glsdisablehyper
\input{sections/acro_list.tex}

\begin{document}

\title{Architectural Optimization and Feature Learning for High-Dimensional Time Series Datasets}

\author{Robert E. Colgan$^{1,2}$, Jingkai Yan$^{2,3}$, Zsuzsa M\'arka$^{4}$, Imre Bartos$^{5}$, Szabolcs M\'arka$^{6}$, and John N. Wright$^{2,3}$}

\address{$^1$Department of Computer Science, Columbia University in the City of New York, 500 W. 120th St., New York, NY 10027, USA\\
$^2$Data Science Institute, Columbia University in the City of New York, 550 W. 120th St., New York, NY 10027, USA\\
$^3$Department of Electrical Engineering, Columbia University in the City of New York, 500 W. 120th St., New York, NY 10027, USA\\
$^4$Columbia Astrophysics Laboratory, Columbia University in the City of New York, 538 W. 120th St., New York, NY 10027, USA\\
$^5$Department of Physics, University of Florida, PO Box 118440, Gainesville, FL 32611-8440, USA\\
$^6$Department of Physics, Columbia University in the City of New York, 538 W. 120th St., New York, NY 10027, USA}

\begin{abstract}
\input{sections/abstract}
\end{abstract}

\maketitle

\input{sections/introduction}

\input{sections/problem_formulation}

\input{sections/learned_features}

\input{sections/sparsity}

\input{sections/deep}

\input{sections/discussion}

\begin{acknowledgments}
\input{sections/acknowledgements}
\end{acknowledgments}

\bibliographystyle{apsrev4-1}
\bibliography{example_paper.bib}

\appendix
\input{sections/appendix}

\end{document}

%% file: sections/acro_list.tex
\newacronym{ligo}{LIGO}{Laser Interferometer Gravitational-wave Observatory}
\newacronym{gw}{GW}{gravitational-wave}
\newacronym{llo}{LLO}{LIGO Livingston Observatory}
\newacronym{er14}{ER14}{LIGO's Engineering Run 14}
\newacronym{o3}{O3}{LIGO's Observing Run 3}
\newacronym{o3b}{O3b}{LIGO's Observing Run 3b}
\newacronym{roc}{ROC}{Receiver Operating Characteristic}
\newacronym{snr}{SNR}{signal-to-noise ratio}
\newacronym{far}{FAR}{false-alarm rate}

\newacronym{sift}{SIFT}{scale-invariant feature transform}
\newacronym{hog}{HOG}{histogram of oriented gradients}
\newacronym{cnn}{CNN}{convolutional neural network}
\newacronym{rnn}{RNN}{recurrent neural network}
\newacronym{lstm}{LSTM}{long short-term memory}
\newacronym{relu}{ReLU}{rectified linear unit}

\newcommand{\mb}{\mathbf}
\renewcommand{\mathbf}{\boldsymbol}
\newcommand{\cmark}{{\color{green}\ding{51}}}%
\newcommand{\xmark}{{\color{red}\ding{55}}}%

\newcommand{\FF}{\texttt{FF}}
\newcommand{\LF}{\texttt{LF}}
\newcommand{\mthree}{\texttt{1Hid}}
\newcommand{\mthreerelu}{\texttt{1HidReLU}}
\newcommand{\vggtwo}{\texttt{VGG6}}
\newcommand{\vggone}{\texttt{VGG13}}
\newcommand{\vggoneBN}{\texttt{VGG13-BN}}
\newcommand{\mr}{\mathrm}
\newcommand{\bb}{\mathbb}

%% file: sections/abstract.tex
As our ability to sense increases, we are experiencing a transition from data-poor problems, in which the central issue is a lack of relevant data, to data-rich problems, in which the central issue is to identify a few relevant features in a sea of observations.
Motivated by applications in gravitational-wave astrophysics, we study a problem in which the goal is to predict the presence of transient noise artifacts in a gravitational wave detector from a rich collection of measurements from the detector and its environment.
We argue that feature learning---in which relevant features are optimized from data---is critical to achieving high accuracy.
We introduce models that reduce the error rate by over 60\% compared to the previous state of the art, which used fixed, hand-crafted features.
Feature learning is useful not only because it can improve performance on prediction tasks; the results provide valuable information about patterns associated with phenomena of interest that would otherwise be impossible to discover.
In our motivating application, features found to be associated with transient noise provide diagnostic information about its origin and suggest mitigation strategies. 
Learning in such a high-dimensional setting is challenging.
Through experiments with a variety of architectures, we identify two key factors in high-performing models: sparsity, for selecting relevant variables within the high-dimensional observations; and depth, which confers flexibility for handling complex interactions and robustness with respect to temporal variations.
We illustrate their significance through a systematic series of experiments on real gravitational-wave detector data.
Our results provide experimental corroboration of common assumptions in the machine-learning community and have direct applicability to improving our ability to sense gravitational waves, as well as to a wide variety of problem settings with similarly high-dimensional, noisy, or partly irrelevant data.

%% file: sections/introduction.tex
\section{Introduction}\label{sec:introduction}

We consider the problem of detecting the presence or absence of some phenomenon of interest from a large collection of time series, a subset of which are predictive, but whose precise mathematical relationship to the phenomenon of interest is a-priori unknown.
Variants of this fundamental problem arise in areas such as finance, neuroscience and brain computer interfaces, structural health monitoring, machine diagnostics, and anomaly detection, just to name a few.
All of these areas present the analyst with time series, which may be noisy, and only a few of which may be relevant to the prediction task at hand. 

Our applied motivation comes from gravitational-wave astrophysics, which uses gravitational phenomena to study the properties of the universe and its occupants.
This scientific quest is driven by extraordinarily sensitive detectors, such as KAGRA, Virgo and LIGO
~\cite{2021PTEP.2021eA101A,2020PhRvD.102f2003B,2020LRR....23....3A,2019PhRvL.123w1107T,2016PhRvD..93k2004M,2016CQGra..33g5009D,2016PhRvL.116m1103A,2015CQGra..32g4001L,2015CQGra..32b4001A,2014CQGra..31v4002A,2013PhRvD..88d3007A,2010CQGra..27h4006H}, which can detect spatial effects as small as ($10^{-19}m/{\sqrt{\mathrm{Hz}}}$).
A major confounding factor is the presence of noise transients, a.k.a.\ \textit{glitches}, in the detector output---non-astrophysical nuisances caused by factors as varied as seismic and ionospheric activity, nearby road and train traffic, optical effects within a detector, etc.
They can obscure or even mimic gravitational waves, so it is important to distinguish them from true astrophysical signals.
One avenue is to make use of terrestrial information about the detector: glitches which can be predicted based on the auxiliary data of the detector alone (without using actual detector output) are unlikely to be astrophysical in origin. 

\cite{EMU} introduced a first demonstration of this concept, leveraging the more than 200,000 additional data streams a detector records (referred to as auxiliary channels) to predict glitches with moderate to high accuracy. 
This method is based on classical tools: one first extracts hand-crafted features from each of the potentially informative subset (about 40,000) of auxiliary channels around the time in question, and then applies sparse logistic regression to perform prediction based on a small subset of these features.

Can we learn better features for glitch detection in gravitational-wave astrophysics?
More generally, under what circumstances is it possible to reliably learn from an overwhelmingly large number of noisy and mostly irrelevant data streams?
Modern machine learning architectures, such as deep neural networks, learn adaptive features from raw data as well as how to combine those features hierarchically.
Compared to classical, manually-defined features, such learned features are better able to capture richer properties of raw data relevant to the task, especially in scenarios with complex and/or high-dimensional data such as natural images and audio.
Moreover, the hierarchical, nonlinear nature of deep neural networks makes them far more powerful than classical linear models, enabling them to learn complex ways of aggregating and synthesizing information from the output of their learned feature detectors \cite{bengio2013representation, krizhevsky2012imagenet, simonyan2014very, Goodfellow-et-al-2016}.

In this work, we systematically explore the properties of several machine learning architectures and evaluate the extent to which they are beneficial for learning a successful classifier in the problem setting described above.
In Sec. \ref{sec:feature_learning} we compare the fixed-feature, logistic regression\textendash based model of \cite{EMU} (which we refer to as \FF{}) to an equivalent linear model with features learned from raw data (which we refer to as \LF{}) and find significant improvement.
In Sec. \ref{sec:sparsity}, we confirm previous findings that regularization-induced sparsity is essential to learning effective classifiers in this setting and extend it to other more complex models.
In Sec. \ref{sec:deep}, we move from flat, linear classifiers to deeper, nonlinear ones and evaluate the effect of depth on performance.
In Sec. \ref{sec:discussion}, we discuss our results and synthesize the observations gleaned from this exploration.
We believe the insights gained will be valuable beyond the problem setting of gravitational-wave astrophysics.

%% file: sections/problem_formulation.tex
\section{Problem Formulation and Prior Work}

We consider the problem of binary classification of a time of interest $t$ from multiple time series.
Given $P$ time series $\mathbf{x}_1, \dots, \mathbf{x}_P$ where each $\mathbf{x}_p$ is a sequence of time-ordered, real-valued scalar samples 
$$\mb x_p = (x_{p,1}, \dots, x_{p,T}) \in \mathbb R^{T}$$ 
sampled at some frequency $f_p$, we would like to make a prediction $y_t \in \{ -1, 1 \}$, where the labels $-1$ and $1$ indicate the presence or absence of some phenomenon of interest at time $t$.
We assume that the $P$ time series (or a subset of them) encode enough information near time $t$ to make such a prediction, but make no further assumptions about the structure or content of the time series.

\subsection{Motivating Application: Glitch Prediction in Gravitational Wave Astrophysics}\label{sec:gw_application}

The above problem formulation is motivated by concerns in gravitational-wave astrophysics, which uses incredibly sensitive interferometric detectors to measure small distortions in spacetime created by astrophysical events such as merging black holes.
In addition to the main gravitational-wave measurement data, the two detectors of the \gls{ligo} project continuously record hundreds of thousands of time series describing a wide array of aspects of the detector's internal and external state and environment.
These are used for monitoring its many components and subsystems to diagnose errors, identify sources of noise, and so on.
In our problem formulation, the time series $\mb x_1, \dots, \mb x_P$ represent these auxiliary measurements. 

One particularly troublesome ongoing phenomenon is the frequent appearance in the main gravitational-wave data stream of brief, loud noise artifacts commonly known within \gls{ligo} as ``glitches."
These glitches, which can appear as frequently as every few seconds, significantly hinder the detectors' sensitivity to the astrophysical phenomena they are intended to measure because they can drown out real astrophysical events or even mimic them, potentially causing false positive detections.
Glitches are thought to be attributable to a wide variety of internal and terrestrial sources; identifying, investigating, and mitigating the various types of glitches that appear and their causes has been a major focus of \gls{ligo}'s engineering efforts for decades \cite{2021CQGra..38s5016S,2021PhRvD.104f2004Y,2021arXiv210812044M,2021CQGra..38m5014D,2021arXiv210707565B,2021CQGra..38n5001N,2021SoftX..1400680C,2020CQGra..37q5001S,2020CQGra..37n5001D,2020arXiv200512761E,2020arXiv200503745C,2018CQGra..35i5016R,2017PhRvD..95j4059M,2017CQGra..34f4003Z,2017PhDT........25V,2016PhDT.......149M,2015CQGra..32x5005N,2013PhRvD..88f2003B,2013PhDT.......555M,2012CQGra..29o5002A,2010CQGra..27s4010C,2010JPhCS.243a2005I,2010JPhCS.243a2006M,2008CQGra..25r4004B,2002nmgm.meet.1841S,gurav2020unsupervised}, often relying on ML techniques.
Currently, many glitch types are identified via methods that directly analyze the gravitational-wave data stream along with one or more auxiliary channels, such as seismic motion monitoring.
Still, many glitches and glitch types are of unknown origin and, if history is a predictor, many new glitch types will emerge in the future. 

In contrast, in this work, our goal is to predict glitches using {\em only} information hidden in undiscovered subsets of the auxiliary channels.
Phenomena which are predictable from only auxiliary channels that are not sensitive to gravitational waves are clearly terrestrial in nature and can be flagged as such.

\paragraph{Ground-truth labels.}
In our general problem formulation, the target labels $y_t \in \{ 1, -1 \}$ represent the presence and absence of glitches in the gravitational-wave data stream, and the goal is to accurately predict these labels.
Following \cite{EMU}, the labels we use for training and evaluation are computed by Omicron \cite{Omicron}, an existing excess power based transient search that directly analyzes the main gravitational-wave data stream to identify glitches.
Also following \cite{EMU}, we choose negative examples (``glitch-free points'') by randomly sampling points in time that are sufficiently distant from any time identified by Omicron as containing a glitch.

\subsection{Prior Work: Glitch Prediction with Fixed Features and Shallow Models}\label{sec:fixed_model}
The possibility of making such predictions was recently demonstrated in the initial work of \cite{EMU}.
The \FF{} method of \cite{EMU} is based on classical statistical tools: it extracts certain {\em hand-crafted} features from the auxiliary channels $\mb x_p$ and then predicts the label $\hat{y}_t$ by linearly combining these features and passing them through a sigmoid function to return a probability estimate: 
    \begin{equation} \label{eqn:lr}
        \hat{y}_t = \sigma\Biggl( \sum_{kp} \omega_{kp} [ \mb f_{k} \star \mb x_p ]_t + b \Biggr).
    \end{equation}
Here, $\star$ denotes discrete correlation and $\sigma(\cdot)$ denotes the logistic function: $\sigma(x) = (1+\exp(-x))^{-1}$.
The filters $\mb f_k$ are fixed; in \cite{EMU}, these correspond to certain intuitively-chosen patterns of behavior that might be predictive, such as spikes and level changes.
The weights $\omega_{kp}$ of this linear combination are learned from training data via gradient descent on an objective function that measures the error between the known ground-truth labels ${y}_t$ and the current model's predictions $\hat{y}_t$.

This method achieves 80-85\% accuracy in glitch detection on unseen validation and test data.
These results demonstrated that glitches can indeed be predicted with moderate accuracy using only auxiliary data and hence that many glitches can be identified as terrestrial in origin and safely discarded, increasing confidence in remaining detection candidates.
While these results were inspiring, the \FF{} method is arguably very far from leveraging all of the structure in these complex datasets, and hence very far from optimal in its ability to predict glitches based on auxiliary channels.
Limitations of this approach include: 
\begin{itemize}
    \item (i) {\bf Hand-crafted vs.\ learned features.} The \FF{} method is based on hand-crafted features designed from intuition-based predictions of a few patterns of behavior that might be predictive; it cannot leverage the ability of modern machine learning techniques to learn more expressive, highly tuned features from raw data \cite{bengio2013representation, Goodfellow-et-al-2016} [Sec. \ref{sec:feature_learning}]
    \item (ii) {\bf Depth.} Increased depth has consistently been found to improve performance and trainability, even over shallower models with equivalent statistical capacity \cite{simonyan2014very, Goodfellow-et-al-2016} [Sec. \ref{sec:deep}]
    \item (iii) {\bf Linear vs.\ nonlinear models.} The ability of modern models to deal with nonlinear structure in data is crucial; deeper hierarchical models without nonlinear activation functions can be reduced to an equivalent flat model \cite{Goodfellow-et-al-2016} [Sec. \ref{sec:deep}]
\end{itemize}
In this paper, we systematically investigate these issues, developing a sequence of models which fundamentally improve over the flat, fixed-feature model discussed above.
We also adopt aspects of that method that prove to be essential to both approaches---most notably regularization-induced sparsity (Sec. \ref{sec:sparsity})---and describe how we adapt them to our proposed methods.

%% file: sections/learned_features.tex
\section{From Fixed to Learned Features}\label{sec:feature_learning}

The principal weakness of the \FF{} method (Eq. \ref{eqn:lr}), as discussed in \cite{EMU} and Sec. \ref{sec:fixed_model}, is that the feature extraction procedure must be defined manually, and optimizing it individually for tens of thousands of time series is not practical.
A major factor in the explosive success of modern machine learning methods in the past decade has been their ability to flexibly learn features from raw data rather than rely on inflexible hand-designed features \cite{bengio2013representation, krizhevsky2012imagenet, Goodfellow-et-al-2016}.
It is natural, then, to consider whether replacing the fixed features of the above model with learned features would improve its performance.

\subsection{Flat Model With Learned Features}\label{sec:lf_model}

\begin{figure*}[t]
\centerline{
            \includegraphics[width=.31\linewidth]{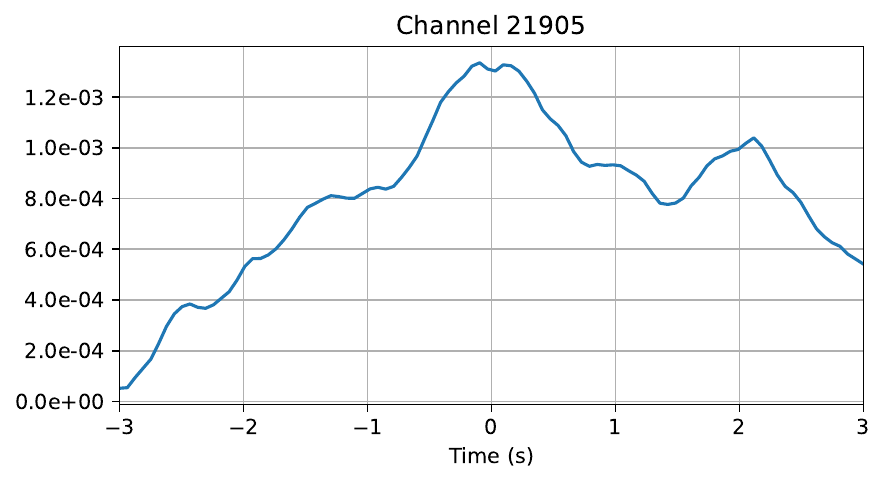}
            \includegraphics[width=.31\linewidth]{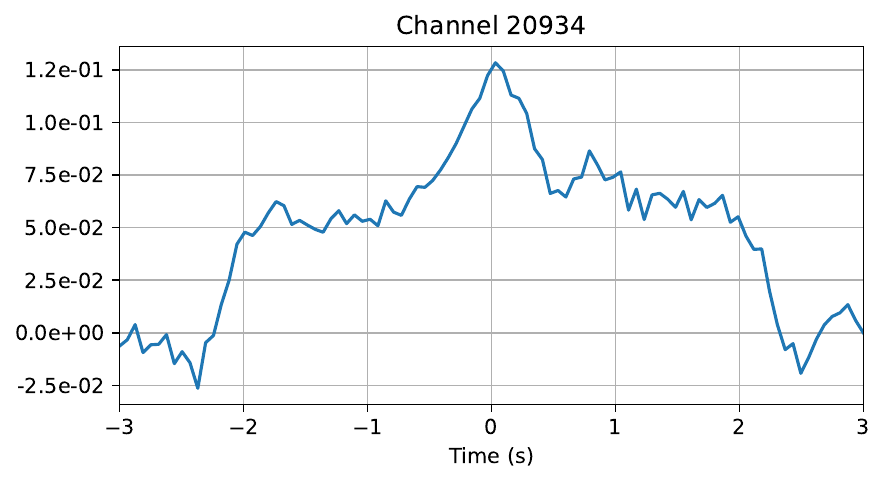}
            \includegraphics[width=.31\linewidth]{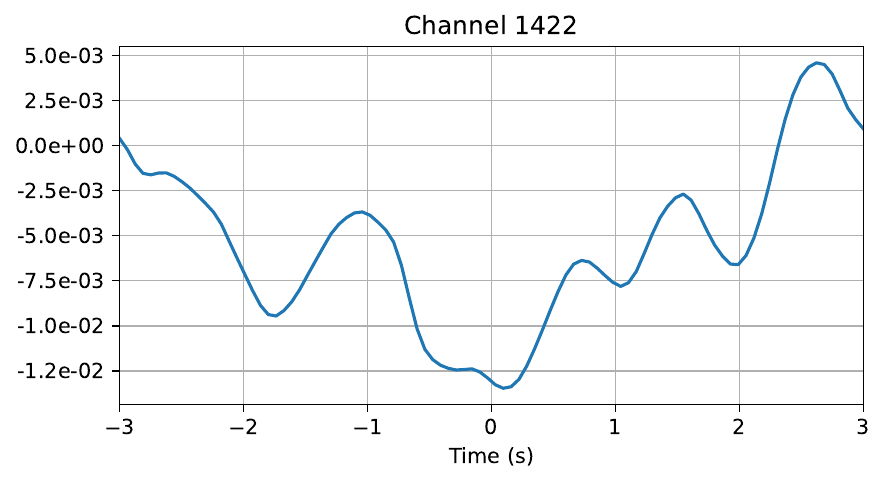}}
\centerline{
            \includegraphics[width=.31\linewidth]{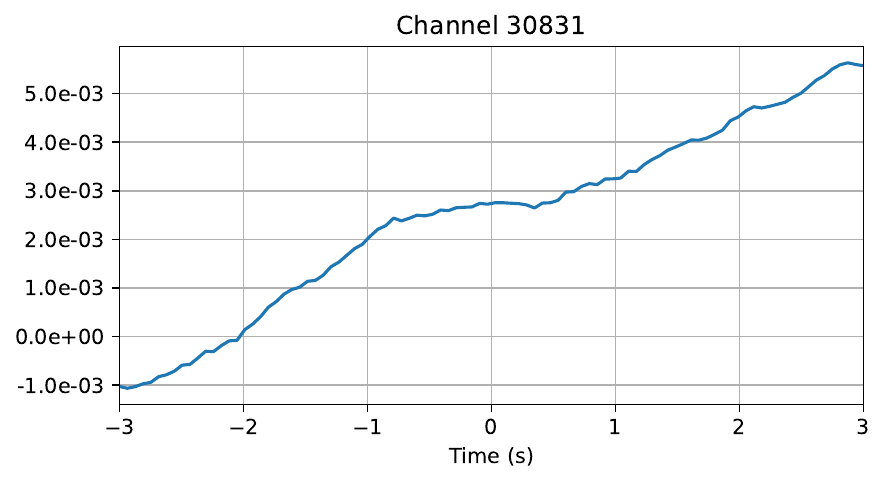}
            \includegraphics[width=.31\linewidth]{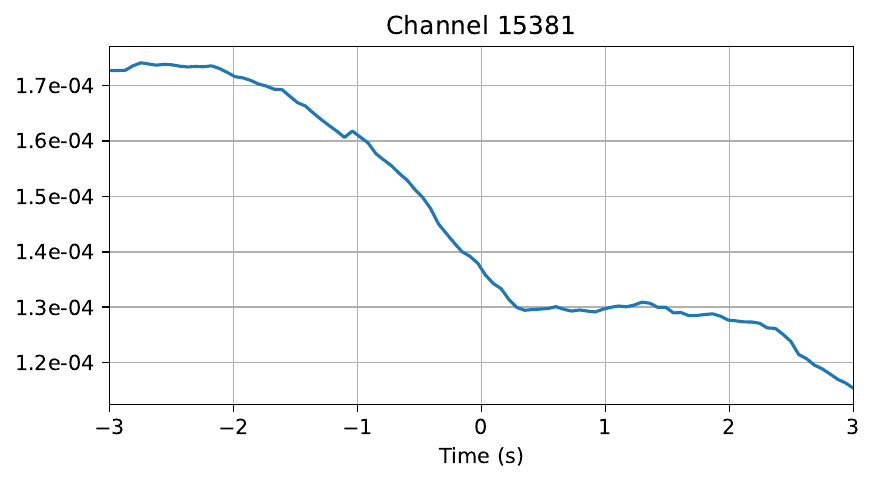}
            \includegraphics[width=.31\linewidth]{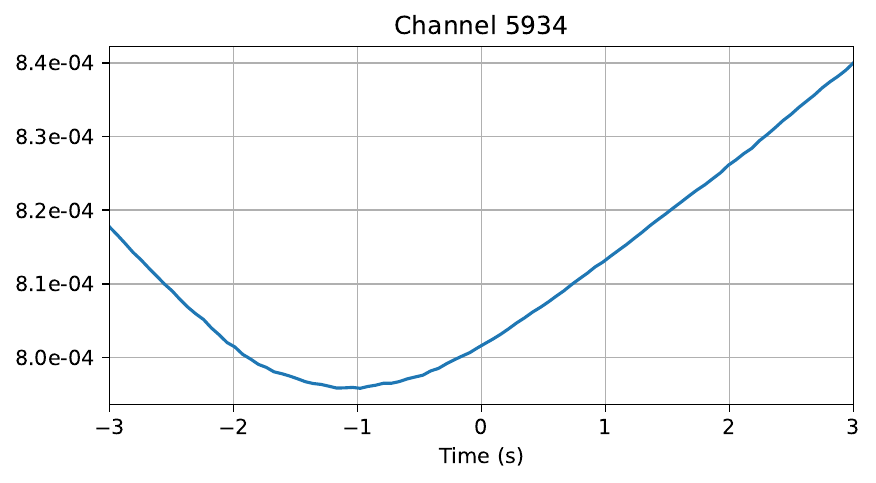}}
\centerline{
            \includegraphics[width=.31\linewidth]{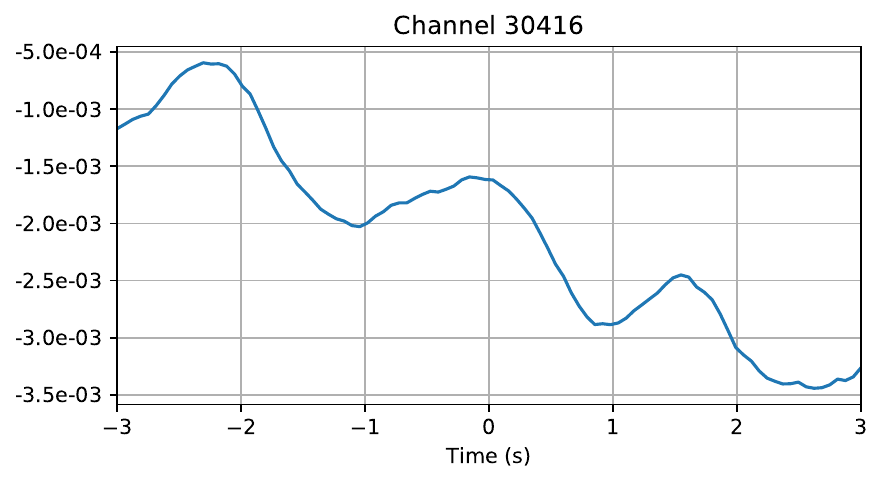}
            \includegraphics[width=.31\linewidth]{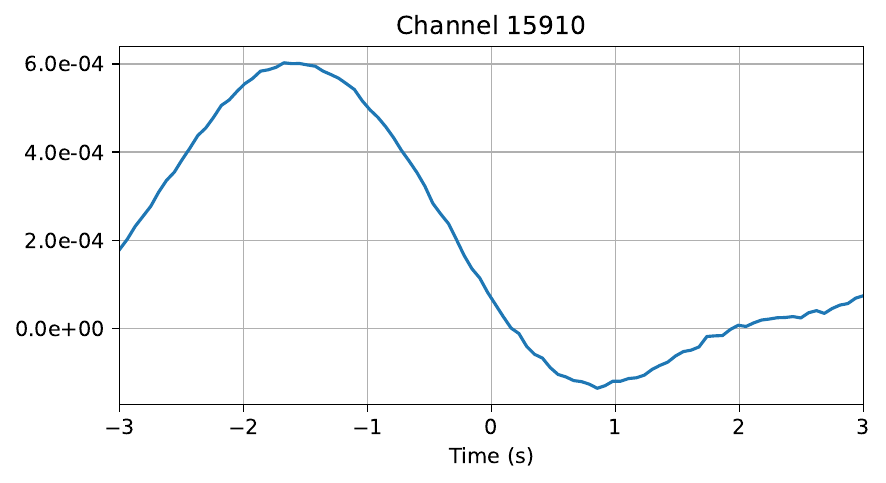}
            \includegraphics[width=.31\linewidth]{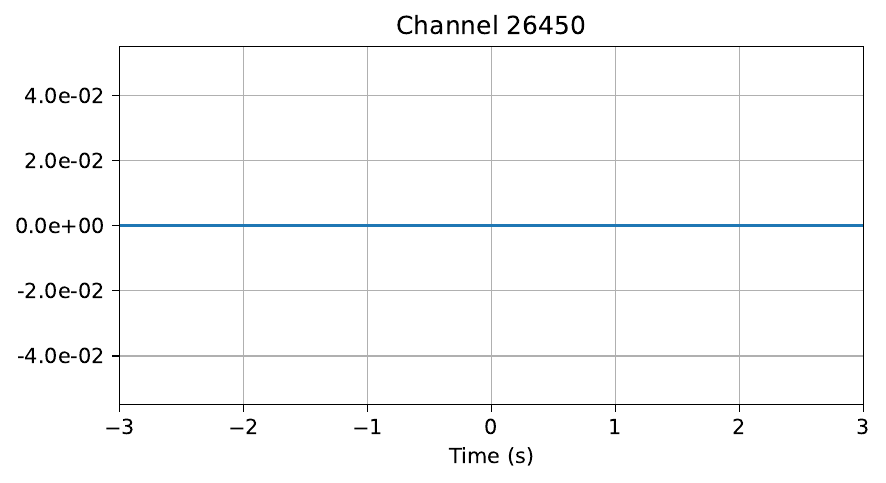}
            }
\caption{A few of the features learned by an \LF{} model with six-second filters.
The x-axes correspond to filter length in seconds; the y-axes show the magnitude of the filter over time.
Intuitively, a higher overall magnitude indicates the associated channel is more important to the model's decisions---when correlated with a higher-magnitude filter, an input data segment will contribute more heavily to the sum and resulting probability estimate than the same segment correlated with a lower-magnitude filter.
The lower-right panel shows a filter with magnitude 0, like the vast majority of the learned filters in the model (see Sec. \ref{sec:sparsity})
}\label{fig:LF_6}
\end{figure*}

To that end, we first consider a nearly equivalent model which differs from \FF{} only in the computation of features from the raw data $\mathbf{x}_p$.
We replace the fixed, manually-defined feature extraction procedure with a convolutional model (defined explicitly in Sec. \ref{sec:conv_notation}) that {\em learns} a filter $\mb w^{0}_{1p}$ ($p \in (1 \dots P)$) for each of the $P$ time series.
For comparison, we preserve for now all other aspects of the \FF{} model, including its linearity and (lack of) depth.

In the general notation of Sec. \ref{sec:conv_notation} (Eq. \ref{eqn:nn-main}), our learned feature model takes the form 
\begin{equation}
    \mb \alpha^1 = \sigma\Bigl( \sum_{p=1}^P \mb w^0_{1p} \star \mb x_p + b \Bigr), 
\end{equation}
where, as in Eq. \ref{eqn:lr}, $\sigma$ is a logistic function.
The estimated probability that $\mb x_t$ belongs to the class ``glitch'' is $\hat{y}_t = \mb \alpha^1_t$.
The filters $\mb w^0_{1p}$ are jointly optimized during the training process.
Below, we refer to this model as \LF{}.
As in the \FF{} model, we also apply a sparsifying regularization term to the filters to encourage $\| \mb w^0_{1p} \|_2 = 0$ for most $p$ (see Section \ref{sec:sparsity}).
The major increase in generality in moving from \FF{} to \LF{} comes from the fact that the $\mb w^0_{1p}$ can be arbitrary vectors---in contrast, \FF{} restricts these filters to be linear combinations $\sum_k \omega_{kp} \mb f_{k}$ of the fixed filters $\mb f_{k}$.\footnote{
The features described by \cite{EMU} include several based on standard deviation, a nonlinear function that cannot be implemented as linear convolution.
Strictly speaking, therefore, it is not correct to say that the \LF{} model is an exact generalization of an \FF{} model that employs standard deviation or other nonlinear functions.
However, the flexibility afforded by learning the feature extractors---even only linear ones---from raw data would almost certainly outweigh any loss of flexibility from restricting ourselves to linear features.
This assumption is consistent with our results with the single-layer \LF{} model.
Such nonlinear functions could be learned by the deeper models discussed in Sec. \ref{sec:deep}.} 

\subsection{Performance Comparison}
We now compare the performance of the \FF{} model with an \LF{} model as described above, setting the input data length to 2.5 s to match the amount of time considered by the \FF{} model for each sample.
(In Sec. \ref{sec:data_length}, we show that longer input lengths enable significantly better performance, further underscoring the advantages and flexibility of learned features.)
For efficiency, we only consider auxiliary channels that are sampled at 16 Hz for both models.

To train the \LF{} and \FF{} models, we follow the procedures described in Section \ref{sec:training} and in \cite{EMU}, respectively, with a few minor modifications to facilitate as direct a comparison as reasonably possible (see Sec. \ref{sec:training} for details).
Following \cite{EMU}, we train and evaluate both models on data from \gls{er14}.
We draw training data from only the final 10,000 s of the 30,000 s training data period because that work found that 10,000 s was a sufficient amount of data for good performance.\footnote{
It is possible that compared to the \FF{} model the models presented here would see greater benefit from a longer training data period because of increased flexibility provided by learned features and other aspects; on the other hand, since the learned features are more closely tuned to the training data, they might be less robust to longer-term changes in the state of the detector, e.g. changes in the shape of glitch-predictive features over time. We leave investigation of the optimal length of time from which to draw training data to future work.}
We do not subsample glitches during this period, using instead all 8,596 glitches and an equal number of glitch-free points (chosen using the procedure described in \cite{EMU}) as training data.
As discussed in Sec. \ref{sec:training}, for the validation results reported we sample a subset of 500 glitches and an equal number of glitch-free points from the validation period; for the test results, we use all glitches present in the test period and an equal number of glitch-free points.

As in \cite{EMU}, for both types of model we perform a grid search over the regularization hyperparameters (using the same grid for both), training models with many parameter settings and evaluating their performance on the validation dataset; we choose the setting that gives the best performance on the validation dataset.

We find that the \FF{} model of \cite{EMU} achieves an accuracy of 85.9\% (with a true positive rate, true negative rate, and loss of 87.6\%, 84.1\%, and 0.3392 respectively)\footnote{
The slightly improved performance of the \FF{} model compared to the same model in \cite{EMU} is most likely due to the combination of modifications to the training procedure described above and the shorter training data period, as \cite{EMU} reported an overall slight decrease in performance as the length of the training period increased beyond 10,000 s.
Also, although we would expect that at least some of the higher-frequency channels contain useful information for the classifier---perhaps even more so, proportionally, than the 16 Hz channels---it is possible that decreasing the data dimensionality improved the model's ability to identify relevant data by enough to outweigh the benefit of the higher-frequency channels.}
on the validation dataset, compared to an accuracy of 87.3\% (TPR 86.1\%, TNR 88.6\%, loss 0.3004) for the \LF{} model, an overall 9.9\% reduction in relative error rate.
On the test dataset, it achieves an accuracy of 85.8\% (TPR 91.2\%, TNR 80.5\%), compared to an accuracy of 88.6\% (TPR 86.7\%, TNR 90.4\%) for the \LF{} model, an overall 19.7\% reduction in relative error rate.

\subsection{Input Segment Length}\label{sec:data_length}

So far, for the sake of comparison with the \FF{} model, we have limited the input data segment length for the \LF{} model to the 2.5 s surrounding each sample time, matching the amount of data used to compute the hand-defined features of the \FF{} model.
As noted in \cite{EMU}, those features---including the length of input data they consider---were chosen largely arbitrarily, and we would like to see whether longer (or shorter) input segments might further improve the performance of learned features.

To that end, we consider input segment length as an additional hyperparameter over which to search while maintaining the same (flat) model structure.
Although it would be possible to implement \FF{} models that accept other segment lengths---and the \FF{} model may well also have benefited from considering longer data segments---it is more straightforward to do so with the \LF{} model: we simply adjust a single hyperparameter and let the model decide how best to make use of the additional data.

\begin{figure}[t]
\centering
\includegraphics[width=\linewidth]{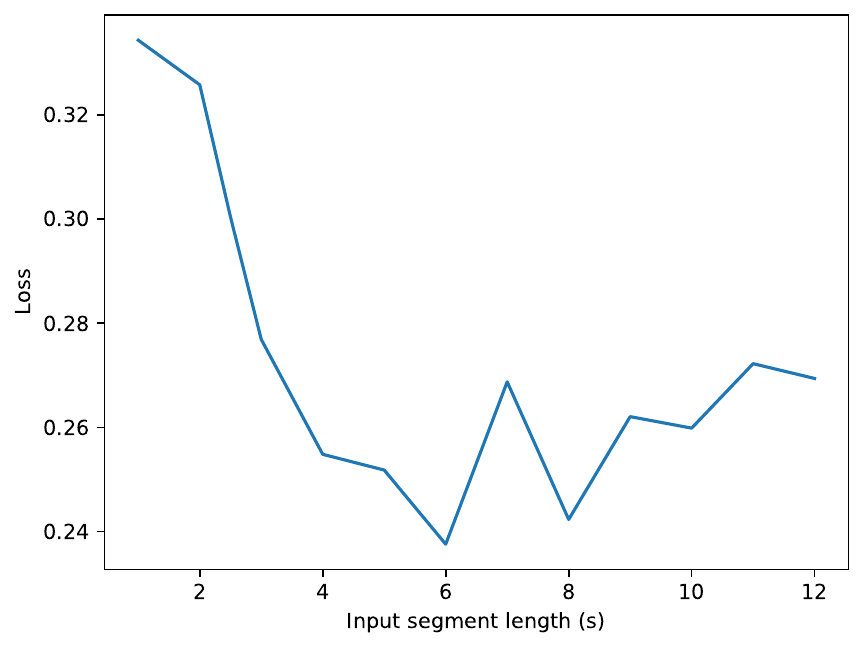}
\caption{Loss on validation dataset (lower is better) vs. feature/input segment length, showing for a given length the best performing model over all hyperparameters tested (i.e., initial learning rate as well as elastic net $\alpha$ and $\lambda$---see Eq. \ref{eqn:elastic} and Fig. \ref{fig:ena_wd_acc})} \label{fig:data_length_s_EN0NoFC}
\end{figure}

Our results indicate that a segment length of four to six seconds is ideal for this model and data (see Fig. \ref{fig:data_length_s_EN0NoFC}), providing significant improvement over the shorter segments considered previously.
Too short, and the model may miss relevant behavior that does not coincide precisely with the appearance of the glitch; too long, and the model may become too difficult to optimize because of the presence of too much extraneous data.

Fig. \ref{fig:LF_6} illustrates some of the features learned in this model.
They can reflect behavior such as local maxima (top left, top center); level changes (center left); oscillatory behavior (bottom center); and more complicated effects specific to each channel that are useful for distinguishing between glitchy and glitch-free times.
Their shapes represent clues to physical/environmental effects that result in glitches and could help diagnose their origins, highlighting an important added benefit of the \LF{} approach.

By accuracy on our validation dataset, the best-performing \LF{} model over every hyperparameter setting tested achieves an accuracy of 90.9\% (TPR 85.8\%, TNR 95.6\%, loss 0.2423).
This represents a 35.5\% reduction in error rate over the \FF{} model discussed above (and a 25.5\% reduction over the \LF{} model with input length limited to 2.5 s).
The lowest validation loss achieved was 0.2376, but this model had slightly worse accuracy at 90.4\% (TPR 91.3\%, TNR 89.5\%).
In Section \ref{sec:deep}, we present experimental results with deeper models that further improve performance.

%% file: sections/sparsity.tex
\section{The Role of Sparsity}\label{sec:sparsity}

A crucial factor in the success of the \FF{} model was the incorporation of a sparsifying regularization term in the optimization objective function---i.e., a term that encourages many of the weights $\omega_{kp}$ to be set to $0$ during training, leaving only the most relevant features to be considered.
Not only did this improve the model's efficiency and interpretability, it also significantly improved its performance compared to a standard, non-sparsifying regularizer on the overall L2 norm of $\mb \omega$. 
The effectiveness of sparse regularization has been observed in a variety of problem settings, leading to widespread adoption of regularizers such as the L1 norm/LASSO \cite{LASSO} and the elastic net \cite{elastic}.
It is particularly relevant in this problem because the vast majority of the $P$ time series contain no useful information for the task.

In developing the \LF{} model, we similarly found sparsity to be an essential property.
Following \cite{EMU}, we employ the elastic net as a regularizer on the magnitudes of the learned filters in the \LF{} model to encourage $\lVert{\mb w^0_{1p} \rVert} = 0$ for most $p$.
The elastic net is a linear combination of L2 and L1 regularization, with tuneable weight on each component:
\begin{equation}\label{eqn:elastic}
    R(\mathbf{\eta}) = \frac{\lambda}{2} \sum_{p} \eta_p^2 + \alpha \sum_{p} |\eta_p|
\end{equation}
where the hyperparameters $\lambda$ and $\alpha$ control the strengths of the L2 and L1 components, respectively.
In our case, we want the regularization to apply to each channel's learned filter $\mb w^0_{1p}$ and act on the filter as a whole, rather than on every sample of all filters independently, so we take $\eta_p = \lVert \mb w^0_{1p} \rVert_2$ in the above equation.
We implement the L1 regularization update following the technique of \cite{tsuruoka2009stochastic}.

\begin{figure}[t]
\centering
\includegraphics[width=\linewidth]{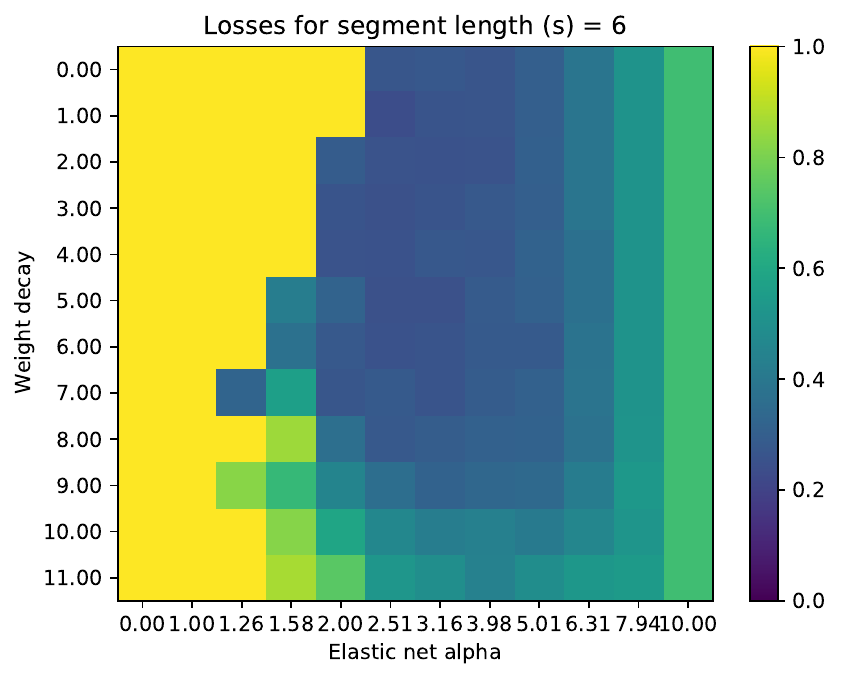}
\caption{Loss on validation dataset (lower is better; values greater than 1 are clipped) as a function of $\alpha$ (x-axis) and $\lambda$ (y-axis) of Eq. \ref{eqn:elastic} for the best-performing initial learning rate.}\label{fig:ena_wd_acc}
\end{figure}

\begin{figure}[t]
\centering
\includegraphics[width=\linewidth]{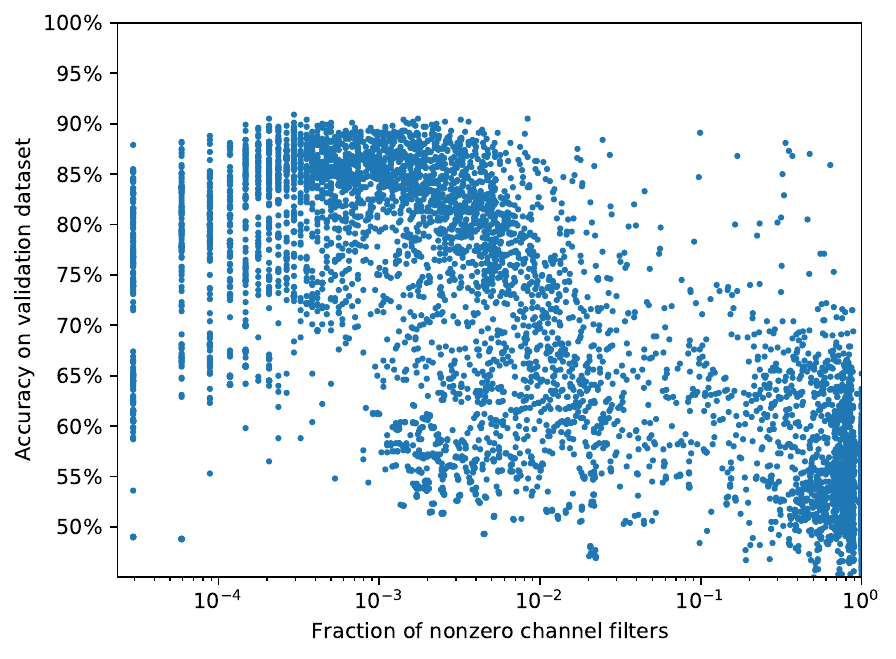}
\caption{Accuracy on validation dataset as a function of the fraction of nonzero channels in the model, for all hyperparameters tested (including filter length).
The best-performing models achieved an accuracy of around 90\% with as few as 0.02\% to 1\% of their approximately 35,000 filters nonzero, corresponding to all but around 10 to a few hundred channels considered irrelevant.} \label{fig:acc_vs_nzchs}
\end{figure}

Following the training procedure described in Sec. \ref{sec:training} with an \LF{} model with no sparsifying regularization, and despite trying a much larger grid of hyperparameter settings, no model at any setting tested was able to achieve more than 64.4\% validation accuracy. 
In contrast, as discussed in Sec. \ref{sec:feature_learning}, sparse models were able to achieve an accuracy of more than 90\% while learning nonzero features for only a small fraction of the 33,939 channels considered (i.e., all but a few channels are ignored by the model when making predictions).
See Fig. \ref{fig:ena_wd_acc} for an illustration of validation accuracy as a function of the sparsity hyperparameters $\alpha$ and $\lambda$ of Eq. \ref{eqn:elastic} and Fig. \ref{fig:acc_vs_nzchs} for an illustration of how validation accuracy correlates with the resulting sparsity of the model.

We also observed that even when we do not explicitly induce sparsity, under certain circumstances training will spontaneously converge to a model that is sparse in one or more respects. 
We describe these findings in Appendix \ref{sec:implicit_sparse}.

%% file: sections/deep.tex
\section{Deeper Models}\label{sec:deep}

\begin{table*}[t]
\input{sections/depth_table}
\end{table*}

In the previous sections, we argued that (i) feature learning and (ii) sparse channel selection are essential ingredients in the design of high-performing glitch predictors. We illustrated these ingredients in the simplest possible setting of shallow (single-layer) architectures. However, experience in application areas such as vision, audio, and natural language processing suggests that feature learning becomes even more powerful in {\em deeper} architectures, which learn hierarchical features. Deeper models have the following potential advantages in glitch prediction: 
\begin{itemize}
    \item {\bf (i) Higher order interactions} between channels are better captured by deep models. A canonical example is the exclusive or relationship, which cannot be represented by a single-layer model. In our setting, this would correspond to the situation in which there are two auxiliary channels which are jointly predictive of a certain type of glitch in the sense that exactly one channel is active (but not both). Deep models are capable of capturing this and other higher order interactions across channels.\footnote{Determining precisely {\em what}, if any, higher order interactions are present across the \gls{ligo} auxiliary channels demands a combination of device modeling and exploratory data analysis. The deeper models proposed in this paper provide one tool for empirically probing the relationship between auxiliary channels and their utility in classifying various glitch types and diagnosing their origins. We will report on this direction in future work.}

    \item {\bf (ii) Robust feature extraction} from individual channels is facilitated by deeper models, in which low-level features are repeatedly combined to produce a hierarchy of increasingly abstract, higher-level features. This robustness is amplified by including {\em pooling} operations at various levels, which increases  robustness to temporal shifts, variations in signal shape, etc.\footnote{Of course, one can also perform pooling in shallow models. Later in this section, we introduce deep models with pooling (\vggtwo{}, \vggone{}, \vggoneBN{}), which achieve state-of-the-art performance on our datasets of interest. The excellent performance of these models should arguably be attributed not just to depth but to the combination of depth, nonlinearity, and pooling.}
    
    \item {\bf (iii) Increased statistical capacity.} Deeper models can accommodate more complicated statistical relationships between the auxiliary channels and the gravitational-wave strain. 
\end{itemize}

In the remainder of this section, we illustrate the power of depth by introducing a sequence of increasingly deeper models, culminating in nonlinear deep models that significantly outperform the previous state-of-the-art for glitch prediction on the datasets considered here.
These results corroborate the utility of depth in feature learning, with the caveat that many of the specific architectures considered here vary in other ways (e.g., presence vs. absence of nonlinearities and temporal pooling).

\begin{figure}[t]
\centering
\includegraphics[width=\linewidth]{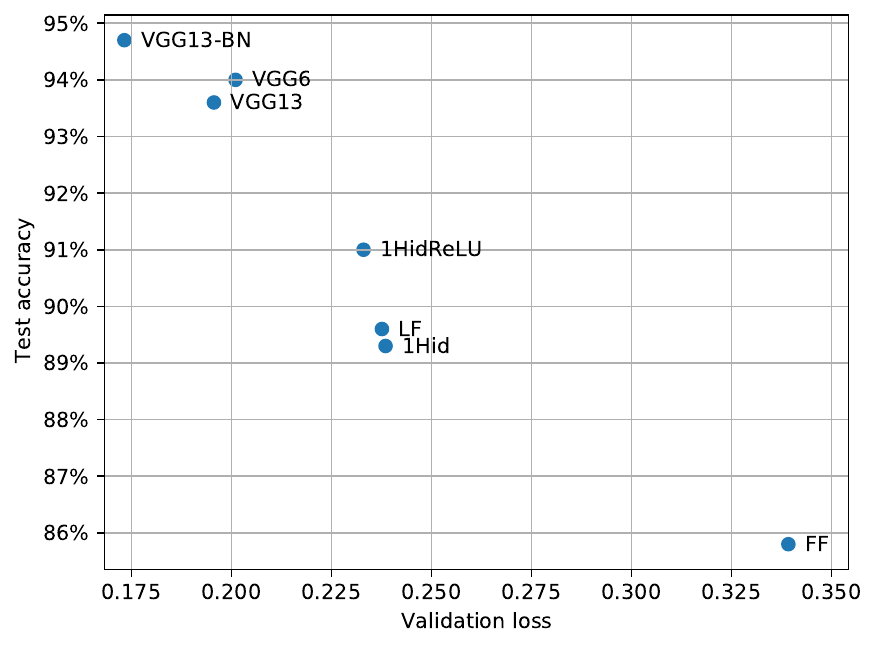}
\caption{Scatter plot of Table \ref{tab:depth-results} showing best validation loss vs. test accuracy for each model.
For validation loss on the x-axis, lower is better, so the models in the upper left have the best overall performance.
Model complexity generally increases from the lower right to the upper left.}
\end{figure}

\paragraph{Models with one hidden layer.}
\mthree{} and \mthreerelu{} both contain a convolutional layer mapping $P$ one-dimensional time-series inputs of length $T$ to a single hidden layer with $P^1$ scalar-valued feature maps.
This is followed by a single fully-connected layer that linearly combines the hidden-layer outputs into a single scalar output and adds a scalar bias; the result is then passed through a sigmoid nonlinearity.
\mthreerelu{} contains a rectifying nonlinearity before the fully-connected layer, whereas \mthree{} does not.
We set $P^1$ to 100 and did not comprehensively study or optimize it, but in limited preliminary experiments we observed little impact from halving or doubling it.
(In fact, as discussed in Sec. \ref{sec:sparse_fmaps}, in many cases the training spontaneously converges to a model with only one or a few active feature maps, but these models can perform as well as or better than models with more active feature maps.)

\paragraph{Models with many hidden layers.}
We also experimented with three deeper models inspired by the VGG16 network \cite{simonyan2014very}, with reduced depth for computational efficiency.
In both models, all convolutional kernels have length three and all max-pooling layers have a kernel size and stride of two.
We fix the length of the input segments to 80 samples (5 seconds) for \vggtwo{} and 200 samples (12.5 seconds) for \vggone{}.
\vggtwo{} consists of a total of five convolutional layers followed by one fully-connected layer, with max-pooling after the second and fifth convolutional layer.
\vggone{} consists of a total of 11 convolutional layers, four max-pooling layers, and two fully-connected layers.
It is identical in structure to \vggtwo{} through the second max-pooling layer, which is followed by two groups of three convolutional layers and a max-pooling layer, then by two fully-connected layers.
\vggoneBN{} is identical to \vggone{} except for the insertion of a batch normalization layer \cite{BatchNorm} before every nonlinearity.

\begin{figure}[t]
\centering
\includegraphics[width=\linewidth]{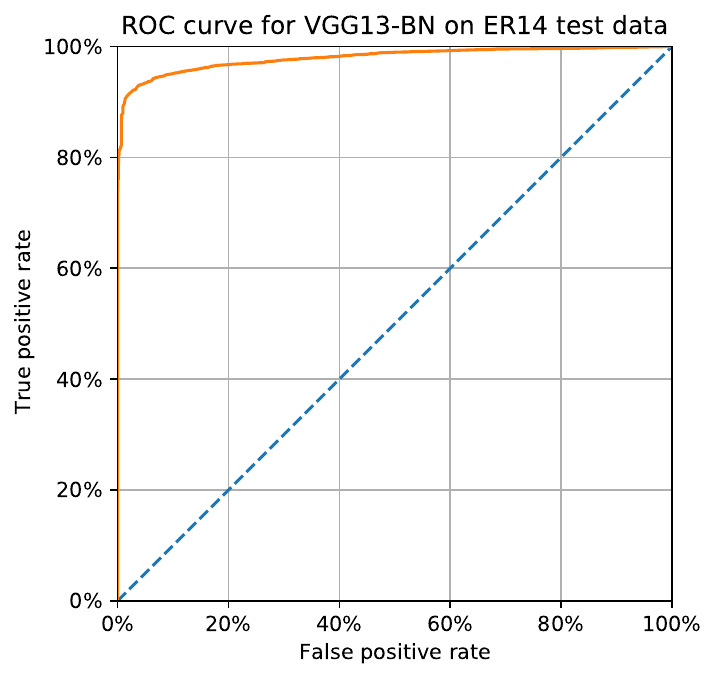}
\caption{ROC curve of \vggoneBN{} on the ER14 test dataset.}\label{fig:vggoneBN_ROC}
\end{figure}

\paragraph{Experimental results with deeper models.} 
We test these deeper models on the same ER14 dataset used in the previous sections.
Table \ref{tab:depth-results} reports the validation accuracy and loss achieved by each model and compares these to the shallow models (\FF{} and \LF{}) introduced in previous sections.
Validation performance increases nearly monotonically with depth; the best-performing model is \vggoneBN{}, which achieves a validation accuracy of 93.1\% and loss of 0.1732 and a test accuracy of 94.7\% and loss of 0.1578.
Fig. \ref{fig:vggoneBN_ROC} shows an ROC curve of \vggoneBN{}'s performance on the test dataset.
It is worth noting that the improved performance in \vggtwo{} and \vggone{} may be attributable not only to their depth but to architectural details such as the use of short convolution filters and pooling.
Nevertheless, Table \ref{tab:depth-results} is consistent with the widely reported finding that deeper networks produce better statistical performance in signal classification tasks.

%% file: sections/depth_table.tex
    \centerline{
    \begin{tabular}{|c|c|c|c|c|c|c|c|}
        \hline
         \bf Model & \bf Feature   & \bf Depth & \bf Nonlinear? & \bf Pooling? & \bf Best Val   & \bf Best Val   & \bf Test Acc \\
                   & \bf Learning? &           &                &              & \bf Loss (Acc) & \bf Acc (Loss) &              \\
         \hline \hline
         \FF{}               & \xmark & 1  & \xmark & \xmark & 0.3392 (85.9\%) & 86.0\% (0.3567) & 85.8\% \\  
         \hline 
         \hline 
         \LF{}               & \cmark & 1  & \xmark & \xmark & 0.2376 (90.4\%) & 90.9\% (0.2423) & 89.6\% \\  
         \hline 
         \mthree{}           & \cmark & 2  & \xmark & \xmark & 0.2385 (90.5\%) & 91.2\% (0.2486) & 89.3\% \\  
         \hline
         \mthreerelu{}       & \cmark & 2  & \cmark & \xmark & 0.2330 (91.0\%) & 91.0\% (0.2330) & 91.0\% \\  
         \hline 
         \vggtwo{}           & \cmark & 6  & \cmark & \cmark & 0.2010 (91.9\%) & 93.0\% (0.2050) & 94.0\% \\  
         \hline 
         \vggone{}           & \cmark & 13 & \cmark & \cmark & 0.1956 (93.4\%) & 93.4\% (0.1956) & 93.6\% \\  
         \hline
         \vggoneBN{}         & \cmark & 13 & \cmark & \cmark & 0.1732 (93.1\%) & 93.6\% (0.1822) & 94.7\% \\  
         \hline
    \end{tabular}
    }
    \caption{Architectural features and performance over the models discussed. Column six shows the best (lowest) loss on the validation dataset over every hyperparameter setting tested for a given model, along with that model's accuracy on the validation dataset. Column seven shows the best accuracy on the validation dataset, along with the loss. To compute accuracy on the test dataset in column eight, we used the model that achieved the best loss.}
    \label{tab:depth-results}

%% file: sections/discussion.tex
\section{Conclusion}\label{sec:discussion}

In this paper, we have demonstrated the potential of feature learning for glitch prediction in gravitational-wave astrophysics and, more generally, for learning from high-dimensional time series.
We have argued that feature learning and architectural choices including sparsity, depth, and nonlinearity are essential to achieving the best possible performance in this setting.
Our architectural explorations culminate in state-of-the-art performance on the ER14 dataset, with a best validation accuracy of 93.6\% and test accuracy of 94.7\% -- an overall approximately 63\% reduction in the test error rate compared to the shallow, fixed feature model!

In general, deeper models require more resources: more training data, and more compute at both training and test time. Obtaining these best possible resource-performance tradeoffs is an important direction for future work; in \cite{YCMMW-new} we study complexity-performance tradeoffs in information extraction from a single time series.
One especially important tradeoff in learning from high-dimensional time series is the tradeoff between sample complexity (how much training data) and test-time performance.
In practice, system characteristics can change over time, and it is important to be able to rapidly adapt to these changes, using limited training data.
Online learning of deep models, using a combination of large offline datasets and limited streaming data, is an important direction for future work.

Feature learning and deep models introduce new opportunities for using machine learning not just as a tool for prediction but as a tool for generating insights into the data generating process.
The models we have described all employ automatic feature learning, which not only improves performance on the classification task compared to fixed features but also can provide valuable diagnostic information---for example, by identifying environmental factors or specific subsystems of a gravitational-wave detector associated with transient noise glitches.
Insight gained during these investigations enable automated adaptability to slowly-changing, time-dependent data as emerging features can be discovered and learned.

Sparse channel selection, as discussed in Section \ref{sec:sparsity}, leads to models that identify a few especially relevant channels for prediction, which---like feature learning---is also beneficial both to performance and interpretability.
Compared to flat, linear, sparse models such as \FF{}, the deep, nonlinear models proposed here squeeze more relevant information out of the small number of selected channels, as witnessed by their substantially improved prediction performance.
Mining these more accurate models for insights into the data generation process is another important direction for future work.

%% file: sections/acknowledgements.tex

We acknowledge computing resources from Columbia University's Shared Research Computing Facility project, which is supported by NIH Research Facility Improvement Grant 1G20RR030893-01, and associated funds from the New York State Empire State Development, Division of Science Technology and Innovation (NYSTAR) Contract C090171, both awarded April 15, 2010.

This material is based upon work and data supported by NSF's LIGO Laboratory which is a major facility fully funded by the National Science Foundation. This research has also made use of data obtained from the Gravitational Wave Open Science Center (\url{https://www.gw-openscience.org}), a  service of LIGO Laboratory, the LIGO Scientific Collaboration and the Virgo Collaboration. LIGO is funded by the U.S. National Science Foundation. Virgo is funded by the French Centre National de Recherche Scientifique (CNRS), the Italian Istituto Nazionale della Fisica Nucleare (INFN) and the Dutch Nikhef, with contributions by Polish and Hungarian institutes.

The authors are grateful for the LIGO Scientific Collaboration review of the paper and this paper is assigned a LIGO DCC number(LIGO-P2200008), with special thanks to Gayathri V. The authors acknowledge the LIGO Collaboration for the production of data used in this study and the LIGO Laboratory for enabling Omicron trigger generation on its computing resources (National Science Foundation Grants PHY-0757058 and PHY-0823459). The authors are grateful to the authors and maintainers of the Omicron and Omega pipelines, the LIGO Commissioning and Detector Characterization Teams and LSC domain expert Colleagues whose fundamental work on the LIGO detectors enabled the data used in this paper. The authors would like to thank colleagues of the LIGO Scientific Collaboration and the Virgo Collaboration for their help and useful comments.

The authors thank the University of Florida and Columbia University in the City of New York for their generous support.
The authors are grateful for the generous support of the National Science Foundation under grant CCF-1740391.
I.B. acknowledges the support of the Alfred P. Sloan Foundation and NSF grants PHY-1911796 and PHY-2110060.


%% file: sections/appendix.tex
\section{Models}

\subsection{Convolutional Model and Notation}\label{sec:conv_notation}
Our models are structured as convolutional neural networks.
A convolutional neural network is comprised of a sequence of $L$ layers, which generate features $\mb \alpha^1, \dots, \mb \alpha^L$.
Each $\mb \alpha^\ell$ consists of $P^\ell$ feature maps (time series) $\mb \alpha^\ell_1, \dots, \mb \alpha^\ell_{P^\ell} \in \mathbb R^{T^\ell}$.
For notational consistency, we let $\mb \alpha^0$ denote the input features $\mb \alpha^0_p = \mb x_p$, with $P^0 = P$ and $T^0 = T$. 

For a given input $\mb x$, features are generated sequentially by applying an affine map followed by (possible) nonlinearity and pooling operations: 
\begin{equation} \label{eqn:nn-main}
\mb \alpha^{\ell+1}_i = \mathcal P^\ell \sigma^\ell \left( \sum_{j=1}^{P^{\ell}} \mb w^{\ell}_{ij} \star \mb \alpha^\ell_j + b^\ell_i \right).
\end{equation}
Here, the $\star$ operation denotes discrete correlation;\footnote{
Correlation is equivalent to a true convolution up to a flipping of the filters $\mb w^\ell_{ij}$. In implementation, the correlation operation may be subsampled (strided) for efficient computation and storage.}
the $\mb w^\ell_{ij} \in \mathbb R^{d^\ell}$ ($i = 1 \dots P^{\ell+1}$, $j = 1 \dots P^{\ell}$) are a collection of filters; and the $b^\ell_i \in \mathbb R$ are scalar biases. 

Because the affine map in Eq. \ref{eqn:nn-main} is built out of correlation operations, it is {\em shift-equivariant}: if $\mathcal{S}_k$ denotes a temporal shift by $k$ samples, 
\begin{equation*}
    \left( \sum_{j=1}^{P^{\ell}} \mb w^{\ell}_{ij} \star \mathcal{S}_k[\mb \alpha^\ell_j] + b^\ell_i \right) = \mathcal{S}_k\left( \sum_{j=1}^{P^{\ell}} \mb w^{\ell}_{ij} \star \mb \alpha^\ell_j + b^\ell_i \right).
\end{equation*}
This is a highly desirable property for analyzing time series.
It ensures that our feature extraction respects the temporal structure of the $\mb \alpha^\ell_j$, and the resulting mapping requires far fewer parameters and far less computation compared to a generic affine map of the same dimension. 

In Eq. \ref{eqn:nn-main}, $\sigma^\ell$ denotes a scalar activation function, which is extended to vector inputs by applying it elementwise.
In our nonlinear models, we often use the ReLU activation $\sigma^\ell(u) = \mathrm{ReLU}(u) = \max\{ u, 0 \}$.
The notation of Eq. \ref{eqn:nn-main} also accommodates ``linear layers'' with no nonlinear activation, simply by setting $\sigma^\ell(u) = u$.
If $\sigma^{\ell}(u) = u$ for all $\ell$, the network output is a linear function of the input $\mb x$.
However, nonlinear models are often preferable due to their greater expressive power: in our experiments, nonlinear models typically outperform their linear counterparts, especially when the number of layers $L$ is large. 

Finally, the operation $\mathcal P^\ell$ performs temporal pooling by taking maxima over contiguous subsets of entries.
This operation is observed to improve robustness to temporal shifts and distortions by aggregating feature responses over time, and it is included in several of the deeper networks that we introduce in Sec. \ref{sec:deep}.
The general notation of Eq. \ref{eqn:nn-main} is flexible enough to accommodate architectures that do not pool simply by taking maxima over subsets consisting of single indices. 

The fixed-feature model \FF{} (Eq. \ref{eqn:lr}) can be seen as an instance of the general model (Eq. \ref{eqn:nn-main}), with $L=1$ layers, with inputs $\mb \alpha^0_p = \mb x_p$, a single output $\hat{\mb y} = \mb \alpha^L$, and filters $\mb w^0_{1p} = \sum_k \omega_{kp} \mb f_{k}$.
That is to say, the logistic predictor is a one-layer neural network, which uses a linear combination of the {\em fixed}, hand-designed features $\mb f_{k}$.

The general model of Eq. \ref{eqn:nn-main} allows for significantly more flexible architectures, in which (i) features can be combined hierarchically, and (ii) features can be learned from data.
In these more flexible architectures, learning is performed in a similar manner to as described above for the \FF{} model---i.e., gradient descent on a measure of the error between the ground-truth labels and the predictions made by the current state of the model predictions; the only major difference is in the greater number of parameters (see Sec. \ref{sec:training} for a detailed description of our training procedures).

\subsection{Model Details}\label{sec:model_descriptions}
In all of the following models, we impose sparsity in the form of elastic net regularizion (as discussed in Sec. \ref{sec:sparsity}) only at the lowest layer, on the connections between the $P$ input channels and the $P^1$ first-layer feature maps.
That is, for a given channel\textendash feature map pair $(p, i)$, the norm of the corresponding learned filter $\lVert{\mb w^0_{ip} \rVert}_2$ corresponds to one element of the vector $\mathbf{\eta}$ on which the elastic net regularization $R(\mathbf{\eta})$ is computed (Eq. \ref{eqn:elastic}).
We implement the L1 component of the elastic net by explicitly computing $\alpha\lVert{\mathbf{\eta}\rVert}_1$ and adding to the loss and the L2 component via standard weight decay, which is applied to every parameter of the model at the same magnitude.

\paragraph{Models with one hidden layer.}
As described in Sec. \ref{sec:deep}, \mthree{} and \mthreerelu{} both contain a convolutional layer mapping $P$ one-dimensional time-series inputs of length $T$ to a single hidden layer with $P^1$ scalar-valued feature maps, followed by a single fully-connected layer that linearly combines the hidden layer outputs into a single scalar-valued output, which is passed through a sigmoid nonlinearity to produce a probability estimate.
The convolutional layer's filters are each the same length $T$ as the input and therefore produce a single scalar value for each pair of an input and a feature map.
As is standard in one-dimensional CNN architectures, for a given feature map $i \in (1 \dots P^1)$, the values obtained from convolving each input $p \in (1 \dots P)$ with the learned filter $\mathbf{w}^0_{ip} \in \mathbb{R}^T$ corresponding to that input\textendash feature map pair are summed and a bias term $b_i$ corresponding to that feature map is added. 
The learned weights for the single convolutional layer therefore consist of a three-way tensor $\mathbf{W} \in \mathbb{R}^{P^1 \times P \times T}$, and the biases consist of a vector $\mathbf{b} \in \mathbb{R}^{P^1}$.

\paragraph{Models with many hidden layers.}
\vggone{}, \vggtwo{}, and \vggoneBN{} are inspired by the VGG16 models of \cite{simonyan2014very} (specifically, configuration D of Table 1), with reduced depth for computational efficiency.
In both models, all convolutional kernels have length three and all max-pooling layers have a kernel size and stride of two.
Each convolutional layer includes a bias for each of its output feature maps, and we employ a rectifying nonlinearity after each convolutional layer.
We do not employ padding, so each convolutional layer outputs a segment two samples shorter than the input and each max-pooling layer halves its input's length.

\vggtwo{} consists of a total of five convolutional layers followed by one fully-connected layer, with max-pooling after the second and fifth convolutional layer.
The first two convolutional layers output 128 feature maps, while the following three output 256.
After the final max-pooling layer, the segment length has been reduced to 16.
With the 256 output feature maps of length 16 as input, the fully-connected layer linearly combines 4,096 inputs into a single scalar-valued output, which is passed through a sigmoid nonlinearity to produce a probability estimate.

\vggone{} consists of a total of 11 convolutional layers, four max-pooling layers, and two fully-connected layers.
It is identical in structure to \vggtwo{} through the second max-pooling layer.
This is followed by two sets of three convolutional layers with 512 feature maps followed by a max-pooling layer.
After the final max-pooling layer, the segment length has been reduced to seven.
With the 512 output feature maps of length seven as input, the first fully-connected layer has 3,584 inputs and 4,096 outputs; the second fully-connected layer linearly combines its 4,096 inputs into a single scalar-valued output, which is passed through a sigmoid nonlinearity to produce a probability estimate.

\section{Datasets and Training Protocols}\label{sec:data_training}

\subsection{Datasets for Glitch Prediction}\label{sec:data}

Following \cite{EMU}, in this paper we consider data from \gls{ligo}'s auxiliary channels during \gls{er14} in March 2019. 
We follow the same procedure to reduce the approximately 250,000 auxiliary channels in a detector to approximately 40,000 by excluding channels that are constant or only vary in a predictable fashion (e.g. counting time cycles). 
Of these, approximately 35,000 have a sample rate of 16 Hz, with the rest having various higher sample rates up to 65,536 Hz; for efficiency, we restrict our analyses in this work to channels with a sample rate of 16 Hz and leave higher-frequency channels to future work.
We further exclude any channels known or suspected to be coupled to the gravitational-wave data stream following the same procedure as \cite{EMU}.

For efficiency, we draw training data from a shorter subset of the ER14 training period used in \cite{EMU} (GPS time 1,235,890,000 to 1,235,900,000, i.e. the final 10,000 seconds of the 30,000-second period of \cite{EMU}) because \cite{EMU} demonstrated that 10,000 seconds is a sufficient amount of time from which to draw training data.
As in \cite{EMU}, we draw validation data from the following 10,000 seconds (GPS time 1,235,900,000 to 1,235,910,000) and test data from the following 10,000 seconds (GPS time 1,235,910,000 to 1,235,920,000).

We normalize each channel by computing the mean and standard deviation of the raw channel data over the entire training data period; then we subtract the training mean and divide by the standard deviation for all data in the training, validation, and test periods.

Following \cite{EMU}, our positive samples are drawn from points in time identified by Omicron \cite{Omicron} as a glitch peak; our negative samples are drawn randomly from periods where no glitch was identified by Omicron within two seconds.
We select the same number of negative samples as there are positive samples in each dataset.

\subsection{Training and Evaluation Protocols}\label{sec:training}

We follow the following training procedure for all models discussed here unless otherwise specified.
As discussed in Sec. \ref{sec:data}, we draw training, validation, and test data from three separate but nearby time periods.
Each model is initialized with the standard Kaiming Uniform \cite{he2015delving} method with the same random seed.
Thereafter, we randomly (again using the same random seed for all models) sample 64 data points from the training period for each training batch and perform stochastic gradient descent with the Adam optimizer \cite{adam}.
The models and training process are implemented in Python with PyTorch \cite{pytorch}.

To evaluate the model during training for learning rate decay and early stopping, we also choose a subset of points from the validation period (the number varies across model types depending on memory constraints, but we use the same points for a given model type).
We evaluate the model on this validation batch every 50 training iterations.
At each validation, if the loss is lower than previously seen, we retain the model state.
When the validation loss fails to decrease for four consecutive validations, we reduce the learning rate by a factor of four; when the validation loss fails to decrease for ten consecutive validations, training ends and we return the model state that performed best on the validation batch.

We also choose a second, larger (1,000-sample) validation batch to evaluate and compare models with different parameter settings.
Once training is complete, we evaluate each model on this larger batch and choose the best-performing one.
The validation accuracies and losses we report are computed for this model on this batch.

Finally, we calculate test accuracy by running only the best-performing model of a given type (chosen based on the second validation batch, as described above) on a dataset consisting of every glitch from the test period and an equal number of appropriately-chosen glitch-free points from the test period.

When directly comparing the \FF{} and \LF{} models, we test several initial learning rates and employ learning rate decay and early stopping based on the loss on a held-out validation subset of the training data rather than running with a fixed learning rate schedule and number of epochs; learning rate decreases by a factor of five after each epoch of no improvement on a validation set until reaching a minimum threshold, at which point training terminates.
We also normalize all data based on the mean and standard deviation of the raw time series over the entire training period rather than normalizing after computing features for the subset of training samples chosen (as was done in \cite{EMU}).
For \LF{}, we also test the same initial learning rates and validate during training with a validation batch as described above every epoch (defined as the model seeing approximately as many samples as present in the training dataset, although not necessarily all of them due to the random sampling procedure used to create the training batches) rather than every 50 training batches. 
Learning rate decreases by a factor of five after each epoch of no improvement until reaching the same minimum threshold as for \FF{}, at which time training terminates.

To determine the best input segment length for \LF{}, as discussed in Sec. \ref{sec:data_length}, we consider it as an additional hyperparameter over which to search while maintaining the same (flat) model structure.
Our results indicate that a segment length of four to six seconds is ideal for this model and data (see Fig. \ref{fig:data_length_s_EN0NoFC}).

\section{Implicit Sparsifying Regularization}\label{sec:implicit_sparse}

In Section \ref{sec:sparsity}, we argued that sparsifying regularization plays a critical role in successful approaches to prediction from large sets of time series: by regularizing the bottom layer weights, we can force the model to select only the most relevant channels, improving both statistical efficiency and interpretability.
We suggested elastic net regularization as a practical and effective means of obtaining sparsity.
Interestingly, even if we do not explicitly apply sparsifying regularization to the network weights, it is still possible to induce sparsity indirectly, through various architectural choices.
In this appendix, we briefly describe two different forms of implicit sparsifying regularization that emerge in certain experiments described in the main body of the paper. 

\subsection{Implicit Regularization for Channel Selection}\label{sec:implicit_sparse_channels}

Our first form of implicit regularization can be motivated through the following experiment, which seems to contradict the claims of Section \ref{sec:sparsity}.
We build a shallow model, in which each input channel is convolved with a channel-specific filter, and then the outputs are linearly combined to produce a final prediction.
We apply weight decay (L2 regularization) to all of the weights of the model, and train in the same manner described in the body of the paper, with an input segment length of 6 seconds.
This approach achieves a validation accuracy of 89.4\%---essentially the same as \LF{}.
(In contrast, as discussed in Sec. \ref{sec:sparsity}, with neither the extra linear layer nor explicit sparse regularization, the best accuracy achieved is 64.4\%.) 
The resulting model is also quite sparse, with all but a few hundred channels having a magnitude of zero or negligibly close to zero.\footnote{
We use ``magnitude'' here to refer to the product of the L2 norm of its learned filter and the corresponding linear weight scalar.
Because of the lack of an explicit approach to handle the nondifferentiability of the implicit sparsifying regularization, most of these magnitudes do not become exactly 0, but the vast majority are smaller than $10^{-9}$.}

There are two surprises here: first, this setup does not involve any explicit sparsifying regularization---just weight decay.
Second, the extra linear layer has no effect on the expressiveness of this model class---because the extra linear layer simply applies a (scalar) linear transform to the output of the first layer, the class of input-output relationships that can be implemented by this two-layer model is exactly the same as that which can be represented by \LF{}.

These surprises are actually linked: one can prove that under L2 regularization, the effect of the extra linear layer is to induce a sparsifying regularization on the first layer filters.
This is essentially a consequence of the basic relationship
\begin{equation}
    \min_{xy = w} \tfrac{1}{2} x^2 + \tfrac{1}{2} y^2 = |w|.
\end{equation}
In words, this says that ``overparameterizing the scalar $w$ by writing it as a product of two quantities $x$ and $y$ that are L2-regularized is equivalent to L1 regularization on $w$.'' 

This phenomenon extends quite broadly.
Let $f : \bb R^d \to \bb R$.
Consider the problem of minimizing $f(\mb w)$ with respect to $\mb w$:
\begin{equation} 
\min_{\mb w \in \bb R^d} f(\mb w) 
\end{equation}
We are interested in what happens if we introduce an additional scalar variable of optimization $\beta$, replace $\mb w$ with $\beta \mb w$ and introduce L2 regularization on both $\mb w$ and $\beta$:
\begin{equation} \label{eqn:extended}
\min_{\mb w \in \bb R^d, \beta} f\Bigl(\beta \mb w\Bigr) + \tfrac{\gamma}{2} \beta^2 + \tfrac{\gamma}{2} \| \mb w \|_2^2. 
\end{equation}
We argue that this extended problem is equivalent to a regularized problem in $\mb w$ only:
\begin{equation}
\min_{\mb w \in \bb R^d} f\Bigl(\mb w \Bigr) + r(\mb w),
\end{equation}
where $r$ is a regularizer.
To that end, consider the following problem 
\begin{equation} 
\min_{\beta \mb w = \mb v} \tfrac{1}{2} \beta^2 + \tfrac{1}{2} \| \mb w \|_2^2
\end{equation}
It is possible to solve this problem in closed form.
$\mb w$ is feasible if and only if $\mb w = s \mb v$ for some $s$.
In this situation, the only feasible $\beta$ is $\beta = \| \mb v \|_2 / \| \mb w \|_2$.
Plugging in, we find an equivalent problem 
\begin{equation}
\min_s  \frac{1}{2 s^2} + \frac{ \| \mb v \|_2^2 s^2 }{2}. 
\end{equation}
Setting the derivative equal to zero, we obtain $s_\star = \frac{1}{ \| \mb v \|_2^{1/2} }$.
Plugging back in, we obtain 
\begin{equation}
\min_{\beta \mb w = \mb v}  \tfrac{1}{2} \beta^2 + \tfrac{1}{2} \| \mb w \|_2^2 = \| \mb v \|_2.
\end{equation}
Applying this observation, our extended problem (Eq. \ref{eqn:extended}) is equivalent to 
\begin{equation}
\min_{\mb w \in \bb R^d} f \Bigl( \mb w \Bigr) +  \gamma \| \mb w \|_2.
\end{equation}
Note that here, the L2 norm of $\mb w$ is not squared.
This is a form of vector sparse regularization which encourages $\mb w = \mb 0$.
These observations can be extended to a multi-filter setting in which there are $K$ vector valued variables of optimization $\mb w_1, \dots, \mb w_K$.
In this setting, adding one extra variable $\beta_i$ for each $\mb w_i$ induces a sum-of-norms regularization:
\begin{align}
&\min_{\mb w_1, \dots, \mb w_K, \beta_1, \dots, \beta_K} f\Bigl( \beta_1 \mb w_1, \beta_2 \mb w_2, \dots, \beta_K \mb w_K \Bigr) \\
& + \sum_{i = 1}^K \tfrac{\gamma}{2} \beta_i^2 + \tfrac{\gamma}{2} \| \mb w_i \|_2^2 \nonumber \\
&\equiv \min_{\mb w_1, \dots, \mb w_K} f\Bigl( \mb w_1, \dots, \mb w_K \Bigr) + \gamma \sum_{i=1}^K \| \mb w_i \|_2. \nonumber
\end{align}
Again, this is a vector sparsity regularizer, which encourages just a few of the $\mb w_i$ to be nonzero.
It is also possible to work out equivalent problems when other regularizers are placed on the auxiliary variables $\beta_i$.
This kind of implicit regularization, in which adding redundant optimization variables dramatically changes the effect of the regularizer, has been demonstrated in a number of previous works (see, e.g., \cite{hoff2017lasso,zhao2019implicit}).

\subsection{Sparsifying Feature Maps with Long Steps}\label{sec:sparse_fmaps}

A different type of sparsification is observed in deeper models with a ReLU nonlinearity: training with a larger learning rate produces more models in which many of the second layer feature maps are identically zero on the entire training dataset, without negatively impacting performance.
In Table \ref{tab:step-results-vggoneBN}, we report both the number of nonzero feature maps and the validation accuracy, for various step sizes $s$.
When $s$ is large, the number of nonzero feature maps can be as small as one.
When $s$ is smaller, the fraction of nonzero feature maps approaches $50\%$.
Interestingly, performance varies only moderately across this range of $s$, even though the nature of the learned model varies significantly.

This phenomenon can be attributed to the ReLU nonlinearity $\sigma(u) = \max\{ u, 0 \}$; its output is identically zero when $u$ is negative.
The composition of the ReLU with an affine function produces a feature $\mb \alpha(\mb x)= \max\{ \mb w^* \mb x + b, 0 \}$ which is identically zero on the halfspace $H_{\mr{off}} = \{ \mb x \mid \mb w^* \mb x + b \le 0 \}$.
If, across the training dataset, all inputs to this function map to this halfspace, this feature will be identically zero.
Moreover, it is likely to stay zero: since 
\begin{equation} 
    \forall \mb x \in \mr{interior}(H_{\mr{off}}), \quad \frac{\partial \alpha(\mb x) }{\partial \mb w} = \mb 0, \; \text{and} \; \frac{\partial \alpha(\mb x) }{\partial b} = 0 \nonumber
\end{equation}
gradient/subgradient updates to $(\mb w, b)$ stay zero.
In the literature, this is sometimes referred to as a {\em dead neuron}. It has been observed both experimentally and theoretically that taking very large steps in $\mb w$ and $b$ tends to push data points $\mb x$ into $H_{\mr{off}}$, producing large numbers of dead neurons, leading to very sparse representations, as reported in Table \ref{tab:step-results-vggoneBN}. 

This type of sparsification may have less overt statistical benefits, although it could convey benefits in terms of interpretability of the learned model and test-time efficiency. 

\begin{table*}[t] 
    \centerline{
    \begin{tabular}{|c|c|c|c|}
        \hline
         \bf Initial step size $s$ & \bf \% of Successful Parameter Settings & \bf Avg. \% Nonzero Feature Maps & \bf Best loss \\
         \hline \hline
            0.001 &  8.9\% & 26.0\% &  0.2881 \\
         \hline 
            0.002 &  14.4\% & 12.2\% & 0.2605 \\
         \hline 
             0.02 &  23.6\% & 9.3\%  & 0.2512 \\
         \hline 
              0.2 &  21.0\% & 2.1\%  & 0.2330 \\
         \hline 
    \end{tabular}
    }
    \caption{In ReLU models, large steps sparsify by producing ``dead neurons''. We trained \mthreerelu{} at several initial learning rates, using the same set of other hyperparameter settings, and evaluated what percentage of these settings performed ``acceptably well," i.e. achieved a loss better than an appropriately chosen threshold (second column). The third column lists the average percentage of nonzero hidden feature map at each learning rate among those well-performing models. The fourth column lists the lowest loss achieved at that learning rate over all other hyperparameters. Interestingly, as initial step size increases, the models become increasingly sparse at the feature map level without sacrificing performance.}
    \label{tab:step-results-m3relu}
\end{table*}

\begin{table*}[t] 
    \centerline{
    \begin{tabular}{|c|c|c|c|}
        \hline
         \bf Initial step size $s$ & \bf \% of Successful Parameter Settings & \bf Avg. \% Nonzero Feature Maps & \bf Best loss \\
         \hline \hline
            0.00025 &  30.2\% & 36.7\% & 0.1956 \\
         \hline 
             0.0005 &  23.3\% & 27.9\% & 0.2032 \\
         \hline 
              0.001 &  9.7\% & 23.8\% &  0.2091 \\
         \hline 
              0.002 &  8.3\% & 17.8\% &  0.2166 \\
         \hline 
    \end{tabular}
    }
    \caption{The same table as above for \vggone. We observe the same behavior as for \mthreerelu{} in the percentage of nonzero feature maps decreasing as the initial learning rate increases. However, although the best loss remains relatively consistent, the percentage of successful parameter settings decreases rather than increases with increasing step size, suggesting that it is more difficult to train deeper models with with higher step sizes.}
    \label{tab:step-results-vggone}
\end{table*}

\begin{table*}[t] 
    \centerline{
    \begin{tabular}{|c|c|c|c|}
        \hline
         \bf Initial step size $s$ & \bf \% of Successful Parameter Settings & \bf Avg. \% Nonzero Feature Maps & \bf Best loss \\
         \hline \hline
            0.00025 &  22.6\% & 15.2\% & 0.1732 \\
         \hline 
             0.0005 &  22.9\% & 20.3\% & 0.1806 \\
         \hline 
              0.001 &  21.9\% & 15.9\% & 0.1822 \\
         \hline 
              0.002 &  17.4\% & 24.6\% & 0.1977 \\
         \hline 
    \end{tabular}
    }
    \caption{The same table as above for \vggoneBN. Batch normalization appears to have a stabilizing effect, reducing the impact of step size on the behaviors observed above.}
    \label{tab:step-results-vggoneBN}
\end{table*}

%% file: example_paper.bbl
\begin{thebibliography}{54}%
\makeatletter
\providecommand \@ifxundefined [1]{%
 \@ifx{#1\undefined}
}%
\providecommand \@ifnum [1]{%
 \ifnum #1\expandafter \@firstoftwo
 \else \expandafter \@secondoftwo
 \fi
}%
\providecommand \@ifx [1]{%
 \ifx #1\expandafter \@firstoftwo
 \else \expandafter \@secondoftwo
 \fi
}%
\providecommand \natexlab [1]{#1}%
\providecommand \enquote  [1]{``#1''}%
\providecommand \bibnamefont  [1]{#1}%
\providecommand \bibfnamefont [1]{#1}%
\providecommand \citenamefont [1]{#1}%
\providecommand \href@noop [0]{\@secondoftwo}%
\providecommand \href [0]{\begingroup \@sanitize@url \@href}%
\providecommand \@href[1]{\@@startlink{#1}\@@href}%
\providecommand \@@href[1]{\endgroup#1\@@endlink}%
\providecommand \@sanitize@url [0]{\catcode `\\12\catcode `\$12\catcode
  `\&12\catcode `\#12\catcode `\^12\catcode `\_12\catcode `\%12\relax}%
\providecommand \@@startlink[1]{}%
\providecommand \@@endlink[0]{}%
\providecommand \url  [0]{\begingroup\@sanitize@url \@url }%
\providecommand \@url [1]{\endgroup\@href {#1}{\urlprefix }}%
\providecommand \urlprefix  [0]{URL }%
\providecommand \Eprint [0]{\href }%
\providecommand \doibase [0]{http://dx.doi.org/}%
\providecommand \selectlanguage [0]{\@gobble}%
\providecommand \bibinfo  [0]{\@secondoftwo}%
\providecommand \bibfield  [0]{\@secondoftwo}%
\providecommand \translation [1]{[#1]}%
\providecommand \BibitemOpen [0]{}%
\providecommand \bibitemStop [0]{}%
\providecommand \bibitemNoStop [0]{.\EOS\space}%
\providecommand \EOS [0]{\spacefactor3000\relax}%
\providecommand \BibitemShut  [1]{\csname bibitem#1\endcsname}%
\let\auto@bib@innerbib\@empty
\bibitem [{\citenamefont {{Akutsu+}}(2021)}]{2021PTEP.2021eA101A}%
  \BibitemOpen
  \bibfield  {author} {\bibinfo {author} {\bibnamefont {{Akutsu+}}},\ }\href
  {\doibase 10.1093/ptep/ptaa125} {\bibfield  {journal} {\bibinfo  {journal}
  {Progress of Theoretical and Experimental Physics}\ }\textbf {\bibinfo
  {volume} {2021}},\ \bibinfo {eid} {05A101} (\bibinfo {year} {2021})},\
  \Eprint {http://arxiv.org/abs/2005.05574} {arXiv:2005.05574
  [physics.ins-det]} \BibitemShut {NoStop}%
\bibitem [{\citenamefont {{Buikema+}}(2020)}]{2020PhRvD.102f2003B}%
  \BibitemOpen
  \bibfield  {author} {\bibinfo {author} {\bibnamefont {{Buikema+}}},\ }\href
  {\doibase 10.1103/PhysRevD.102.062003} {\bibfield  {journal} {\bibinfo
  {journal} {\prd}\ }\textbf {\bibinfo {volume} {102}},\ \bibinfo {eid}
  {062003} (\bibinfo {year} {2020})},\ \Eprint
  {http://arxiv.org/abs/2008.01301} {arXiv:2008.01301 [astro-ph.IM]}
  \BibitemShut {NoStop}%
\bibitem [{\citenamefont {{Abbott+}}(2020)}]{2020LRR....23....3A}%
  \BibitemOpen
  \bibfield  {author} {\bibinfo {author} {\bibnamefont {{Abbott+}}},\ }\href
  {\doibase 10.1007/s41114-020-00026-9} {\bibfield  {journal} {\bibinfo
  {journal} {Living Reviews in Relativity}\ }\textbf {\bibinfo {volume} {23}},\
  \bibinfo {eid} {3} (\bibinfo {year} {2020})}\BibitemShut {NoStop}%
\bibitem [{\citenamefont {{Tse+}}(2019)}]{2019PhRvL.123w1107T}%
  \BibitemOpen
  \bibfield  {author} {\bibinfo {author} {\bibnamefont {{Tse+}}},\ }\href
  {\doibase 10.1103/PhysRevLett.123.231107} {\bibfield  {journal} {\bibinfo
  {journal} {\prl}\ }\textbf {\bibinfo {volume} {123}},\ \bibinfo {eid}
  {231107} (\bibinfo {year} {2019})}\BibitemShut {NoStop}%
\bibitem [{\citenamefont {{Martynov+}}(2016)}]{2016PhRvD..93k2004M}%
  \BibitemOpen
  \bibfield  {author} {\bibinfo {author} {\bibnamefont {{Martynov+}}},\ }\href
  {\doibase 10.1103/PhysRevD.93.112004} {\bibfield  {journal} {\bibinfo
  {journal} {\prd}\ }\textbf {\bibinfo {volume} {93}},\ \bibinfo {eid} {112004}
  (\bibinfo {year} {2016})},\ \Eprint {http://arxiv.org/abs/1604.00439}
  {arXiv:1604.00439 [astro-ph.IM]} \BibitemShut {NoStop}%
\bibitem [{\citenamefont {{Dooley+}}(2016)}]{2016CQGra..33g5009D}%
  \BibitemOpen
  \bibfield  {author} {\bibinfo {author} {\bibnamefont {{Dooley+}}},\ }\href
  {\doibase 10.1088/0264-9381/33/7/075009} {\bibfield  {journal} {\bibinfo
  {journal} {Classical and Quantum Gravity}\ }\textbf {\bibinfo {volume}
  {33}},\ \bibinfo {eid} {075009} (\bibinfo {year} {2016})},\ \Eprint
  {http://arxiv.org/abs/1510.00317} {arXiv:1510.00317 [physics.ins-det]}
  \BibitemShut {NoStop}%
\bibitem [{\citenamefont {{Abbott++}}(2016)}]{2016PhRvL.116m1103A}%
  \BibitemOpen
  \bibfield  {author} {\bibinfo {author} {\bibnamefont {{Abbott++}}},\ }\href
  {\doibase 10.1103/PhysRevLett.116.131103} {\bibfield  {journal} {\bibinfo
  {journal} {\prl}\ }\textbf {\bibinfo {volume} {116}},\ \bibinfo {eid}
  {131103} (\bibinfo {year} {2016})},\ \Eprint
  {http://arxiv.org/abs/1602.03838} {arXiv:1602.03838 [gr-qc]} \BibitemShut
  {NoStop}%
\bibitem [{\citenamefont {Aasi}\ \emph {et~al.}(2015)\citenamefont {Aasi},
  \citenamefont {Abbott}, \citenamefont {Abbott}, \citenamefont {Abbott},
  \citenamefont {Abernathy}, \citenamefont {Ackley}, \citenamefont {Adams},
  \citenamefont {Adams}, \citenamefont {Addesso}, \citenamefont {Adhikari}
  \emph {et~al.}}]{2015CQGra..32g4001L}%
  \BibitemOpen
  \bibfield  {author} {\bibinfo {author} {\bibfnamefont {J.}~\bibnamefont
  {Aasi}}, \bibinfo {author} {\bibfnamefont {B.}~\bibnamefont {Abbott}},
  \bibinfo {author} {\bibfnamefont {R.}~\bibnamefont {Abbott}}, \bibinfo
  {author} {\bibfnamefont {T.}~\bibnamefont {Abbott}}, \bibinfo {author}
  {\bibfnamefont {M.}~\bibnamefont {Abernathy}}, \bibinfo {author}
  {\bibfnamefont {K.}~\bibnamefont {Ackley}}, \bibinfo {author} {\bibfnamefont
  {C.}~\bibnamefont {Adams}}, \bibinfo {author} {\bibfnamefont
  {T.}~\bibnamefont {Adams}}, \bibinfo {author} {\bibfnamefont
  {P.}~\bibnamefont {Addesso}}, \bibinfo {author} {\bibfnamefont
  {R.}~\bibnamefont {Adhikari}},  \emph {et~al.},\ }\href {\doibase
  10.1088/0264-9381/32/7/074001} {\bibfield  {journal} {\bibinfo  {journal}
  {Classical and Quantum Gravity}\ }\textbf {\bibinfo {volume} {32}},\ \bibinfo
  {eid} {074001} (\bibinfo {year} {2015})},\ \Eprint
  {http://arxiv.org/abs/1411.4547} {arXiv:1411.4547 [gr-qc]} \BibitemShut
  {NoStop}%
\bibitem [{\citenamefont {Acernese}\ \emph {et~al.}(2015)\citenamefont
  {Acernese}, \citenamefont {Agathos}, \citenamefont {Agatsuma}, \citenamefont
  {Aisa}, \citenamefont {Allemandou}, \citenamefont {Allocca}, \citenamefont
  {Amarni}, \citenamefont {Astone}, \citenamefont {Balestri}, \citenamefont
  {Ballardin} \emph {et~al.}}]{2015CQGra..32b4001A}%
  \BibitemOpen
  \bibfield  {author} {\bibinfo {author} {\bibfnamefont {F.~a.}\ \bibnamefont
  {Acernese}}, \bibinfo {author} {\bibfnamefont {M.}~\bibnamefont {Agathos}},
  \bibinfo {author} {\bibfnamefont {K.}~\bibnamefont {Agatsuma}}, \bibinfo
  {author} {\bibfnamefont {D.}~\bibnamefont {Aisa}}, \bibinfo {author}
  {\bibfnamefont {N.}~\bibnamefont {Allemandou}}, \bibinfo {author}
  {\bibfnamefont {A.}~\bibnamefont {Allocca}}, \bibinfo {author} {\bibfnamefont
  {J.}~\bibnamefont {Amarni}}, \bibinfo {author} {\bibfnamefont
  {P.}~\bibnamefont {Astone}}, \bibinfo {author} {\bibfnamefont
  {G.}~\bibnamefont {Balestri}}, \bibinfo {author} {\bibfnamefont
  {G.}~\bibnamefont {Ballardin}},  \emph {et~al.},\ }\href {\doibase
  10.1088/0264-9381/32/2/024001} {\bibfield  {journal} {\bibinfo  {journal}
  {Classical and Quantum Gravity}\ }\textbf {\bibinfo {volume} {32}},\ \bibinfo
  {eid} {024001} (\bibinfo {year} {2015})},\ \Eprint
  {http://arxiv.org/abs/1408.3978} {arXiv:1408.3978 [gr-qc]} \BibitemShut
  {NoStop}%
\bibitem [{\citenamefont {{Affeldt+}}(2014)}]{2014CQGra..31v4002A}%
  \BibitemOpen
  \bibfield  {author} {\bibinfo {author} {\bibnamefont {{Affeldt+}}},\ }\href
  {\doibase 10.1088/0264-9381/31/22/224002} {\bibfield  {journal} {\bibinfo
  {journal} {Classical and Quantum Gravity}\ }\textbf {\bibinfo {volume}
  {31}},\ \bibinfo {eid} {224002} (\bibinfo {year} {2014})}\BibitemShut
  {NoStop}%
\bibitem [{\citenamefont {{Aso+}}(2013)}]{2013PhRvD..88d3007A}%
  \BibitemOpen
  \bibfield  {author} {\bibinfo {author} {\bibnamefont {{Aso+}}},\ }\href
  {\doibase 10.1103/PhysRevD.88.043007} {\bibfield  {journal} {\bibinfo
  {journal} {\prd}\ }\textbf {\bibinfo {volume} {88}},\ \bibinfo {eid} {043007}
  (\bibinfo {year} {2013})},\ \Eprint {http://arxiv.org/abs/1306.6747}
  {arXiv:1306.6747 [gr-qc]} \BibitemShut {NoStop}%
\bibitem [{\citenamefont {Harry+}(2010)}]{2010CQGra..27h4006H}%
  \BibitemOpen
  \bibfield  {author} {\bibinfo {author} {\bibnamefont {Harry+}},\ }\href
  {\doibase 10.1088/0264-9381/27/8/084006} {\bibfield  {journal} {\bibinfo
  {journal} {Classical and Quantum Gravity}\ }\textbf {\bibinfo {volume}
  {27}},\ \bibinfo {eid} {084006} (\bibinfo {year} {2010})}\BibitemShut
  {NoStop}%
\bibitem [{\citenamefont {Colgan}\ \emph {et~al.}(2020)\citenamefont {Colgan},
  \citenamefont {Corley}, \citenamefont {Lau}, \citenamefont {Bartos},
  \citenamefont {Wright}, \citenamefont {M\'arka},\ and\ \citenamefont
  {M\'arka}}]{EMU}%
  \BibitemOpen
  \bibfield  {author} {\bibinfo {author} {\bibfnamefont {R.~E.}\ \bibnamefont
  {Colgan}}, \bibinfo {author} {\bibfnamefont {K.~R.}\ \bibnamefont {Corley}},
  \bibinfo {author} {\bibfnamefont {Y.}~\bibnamefont {Lau}}, \bibinfo {author}
  {\bibfnamefont {I.}~\bibnamefont {Bartos}}, \bibinfo {author} {\bibfnamefont
  {J.~N.}\ \bibnamefont {Wright}}, \bibinfo {author} {\bibfnamefont
  {Z.}~\bibnamefont {M\'arka}}, \ and\ \bibinfo {author} {\bibfnamefont
  {S.}~\bibnamefont {M\'arka}},\ }\href {\doibase 10.1103/PhysRevD.101.102003}
  {\bibfield  {journal} {\bibinfo  {journal} {Phys. Rev. D}\ }\textbf {\bibinfo
  {volume} {101}},\ \bibinfo {pages} {102003} (\bibinfo {year}
  {2020})}\BibitemShut {NoStop}%
\bibitem [{\citenamefont {Bengio}\ \emph {et~al.}(2013)\citenamefont {Bengio},
  \citenamefont {Courville},\ and\ \citenamefont
  {Vincent}}]{bengio2013representation}%
  \BibitemOpen
  \bibfield  {author} {\bibinfo {author} {\bibfnamefont {Y.}~\bibnamefont
  {Bengio}}, \bibinfo {author} {\bibfnamefont {A.}~\bibnamefont {Courville}}, \
  and\ \bibinfo {author} {\bibfnamefont {P.}~\bibnamefont {Vincent}},\
  }\href@noop {} {\bibfield  {journal} {\bibinfo  {journal} {IEEE transactions
  on pattern analysis and machine intelligence}\ }\textbf {\bibinfo {volume}
  {35}},\ \bibinfo {pages} {1798} (\bibinfo {year} {2013})}\BibitemShut
  {NoStop}%
\bibitem [{\citenamefont {Krizhevsky}\ \emph {et~al.}(2012)\citenamefont
  {Krizhevsky}, \citenamefont {Sutskever},\ and\ \citenamefont
  {Hinton}}]{krizhevsky2012imagenet}%
  \BibitemOpen
  \bibfield  {author} {\bibinfo {author} {\bibfnamefont {A.}~\bibnamefont
  {Krizhevsky}}, \bibinfo {author} {\bibfnamefont {I.}~\bibnamefont
  {Sutskever}}, \ and\ \bibinfo {author} {\bibfnamefont {G.~E.}\ \bibnamefont
  {Hinton}},\ }in\ \href@noop {} {\emph {\bibinfo {booktitle} {Advances in
  neural information processing systems}}}\ (\bibinfo {year} {2012})\ pp.\
  \bibinfo {pages} {1097--1105}\BibitemShut {NoStop}%
\bibitem [{\citenamefont {Simonyan}\ and\ \citenamefont
  {Zisserman}(2014)}]{simonyan2014very}%
  \BibitemOpen
  \bibfield  {author} {\bibinfo {author} {\bibfnamefont {K.}~\bibnamefont
  {Simonyan}}\ and\ \bibinfo {author} {\bibfnamefont {A.}~\bibnamefont
  {Zisserman}},\ }\href@noop {} {\bibfield  {journal} {\bibinfo  {journal}
  {arXiv preprint arXiv:1409.1556}\ } (\bibinfo {year} {2014})}\BibitemShut
  {NoStop}%
\bibitem [{\citenamefont {Goodfellow}\ \emph {et~al.}(2016)\citenamefont
  {Goodfellow}, \citenamefont {Bengio},\ and\ \citenamefont
  {Courville}}]{Goodfellow-et-al-2016}%
  \BibitemOpen
  \bibfield  {author} {\bibinfo {author} {\bibfnamefont {I.}~\bibnamefont
  {Goodfellow}}, \bibinfo {author} {\bibfnamefont {Y.}~\bibnamefont {Bengio}},
  \ and\ \bibinfo {author} {\bibfnamefont {A.}~\bibnamefont {Courville}},\
  }\href@noop {} {\emph {\bibinfo {title} {Deep Learning}}}\ (\bibinfo
  {publisher} {MIT Press},\ \bibinfo {year} {2016})\ \bibinfo {note}
  {\url{http://www.deeplearningbook.org}}\BibitemShut {NoStop}%
\bibitem [{\citenamefont {{Soni}}\ \emph {et~al.}(2021)\citenamefont {{Soni}},
  \citenamefont {{Berry}}, \citenamefont {{Coughlin}}, \citenamefont
  {{Harandi}}, \citenamefont {{Jackson}}, \citenamefont {{Crowston}},
  \citenamefont {{{\O}sterlund}}, \citenamefont {{Patane}}, \citenamefont
  {{Katsaggelos}}, \citenamefont {{Trouille}}, \citenamefont {{Baranowski}},
  \citenamefont {{Domainko}}, \citenamefont {{Kaminski}}, \citenamefont
  {{Rodriguez}}, \citenamefont {{Marciniak}}, \citenamefont {{Nauta}},
  \citenamefont {{Niklasch}}, \citenamefont {{Rote}}, \citenamefont
  {{T{\'e}gl{\'a}s}}, \citenamefont {{Unsworth}},\ and\ \citenamefont
  {{Zhang}}}]{2021CQGra..38s5016S}%
  \BibitemOpen
  \bibfield  {author} {\bibinfo {author} {\bibfnamefont {S.}~\bibnamefont
  {{Soni}}}, \bibinfo {author} {\bibfnamefont {C.~P.~L.}\ \bibnamefont
  {{Berry}}}, \bibinfo {author} {\bibfnamefont {S.~B.}\ \bibnamefont
  {{Coughlin}}}, \bibinfo {author} {\bibfnamefont {M.}~\bibnamefont
  {{Harandi}}}, \bibinfo {author} {\bibfnamefont {C.~B.}\ \bibnamefont
  {{Jackson}}}, \bibinfo {author} {\bibfnamefont {K.}~\bibnamefont
  {{Crowston}}}, \bibinfo {author} {\bibfnamefont {C.}~\bibnamefont
  {{{\O}sterlund}}}, \bibinfo {author} {\bibfnamefont {O.}~\bibnamefont
  {{Patane}}}, \bibinfo {author} {\bibfnamefont {A.~K.}\ \bibnamefont
  {{Katsaggelos}}}, \bibinfo {author} {\bibfnamefont {L.}~\bibnamefont
  {{Trouille}}}, \bibinfo {author} {\bibfnamefont {V.~G.}\ \bibnamefont
  {{Baranowski}}}, \bibinfo {author} {\bibfnamefont {W.~F.}\ \bibnamefont
  {{Domainko}}}, \bibinfo {author} {\bibfnamefont {K.}~\bibnamefont
  {{Kaminski}}}, \bibinfo {author} {\bibfnamefont {M.~A.~L.}\ \bibnamefont
  {{Rodriguez}}}, \bibinfo {author} {\bibfnamefont {U.}~\bibnamefont
  {{Marciniak}}}, \bibinfo {author} {\bibfnamefont {P.}~\bibnamefont
  {{Nauta}}}, \bibinfo {author} {\bibfnamefont {G.}~\bibnamefont {{Niklasch}}},
  \bibinfo {author} {\bibfnamefont {R.~R.}\ \bibnamefont {{Rote}}}, \bibinfo
  {author} {\bibfnamefont {B.}~\bibnamefont {{T{\'e}gl{\'a}s}}}, \bibinfo
  {author} {\bibfnamefont {C.}~\bibnamefont {{Unsworth}}}, \ and\ \bibinfo
  {author} {\bibfnamefont {C.}~\bibnamefont {{Zhang}}},\ }\href {\doibase
  10.1088/1361-6382/ac1ccb} {\bibfield  {journal} {\bibinfo  {journal}
  {Classical and Quantum Gravity}\ }\textbf {\bibinfo {volume} {38}},\ \bibinfo
  {eid} {195016} (\bibinfo {year} {2021})},\ \Eprint
  {http://arxiv.org/abs/2103.12104} {arXiv:2103.12104 [gr-qc]} \BibitemShut
  {NoStop}%
\bibitem [{\citenamefont {{Yu}}\ \emph {et~al.}(2021)\citenamefont {{Yu}},
  \citenamefont {{Adhikari}}, \citenamefont {{Magee}}, \citenamefont
  {{Sachdev}},\ and\ \citenamefont {{Chen}}}]{2021PhRvD.104f2004Y}%
  \BibitemOpen
  \bibfield  {author} {\bibinfo {author} {\bibfnamefont {H.}~\bibnamefont
  {{Yu}}}, \bibinfo {author} {\bibfnamefont {R.~X.}\ \bibnamefont
  {{Adhikari}}}, \bibinfo {author} {\bibfnamefont {R.}~\bibnamefont {{Magee}}},
  \bibinfo {author} {\bibfnamefont {S.}~\bibnamefont {{Sachdev}}}, \ and\
  \bibinfo {author} {\bibfnamefont {Y.}~\bibnamefont {{Chen}}},\ }\href
  {\doibase 10.1103/PhysRevD.104.062004} {\bibfield  {journal} {\bibinfo
  {journal} {\prd}\ }\textbf {\bibinfo {volume} {104}},\ \bibinfo {eid}
  {062004} (\bibinfo {year} {2021})},\ \Eprint
  {http://arxiv.org/abs/2104.09438} {arXiv:2104.09438 [gr-qc]} \BibitemShut
  {NoStop}%
\bibitem [{\citenamefont {{Merritt}}\ \emph {et~al.}(2021)\citenamefont
  {{Merritt}}, \citenamefont {{Farr}}, \citenamefont {{Hur}}, \citenamefont
  {{Edelman}},\ and\ \citenamefont {{Doctor}}}]{2021arXiv210812044M}%
  \BibitemOpen
  \bibfield  {author} {\bibinfo {author} {\bibfnamefont {J.}~\bibnamefont
  {{Merritt}}}, \bibinfo {author} {\bibfnamefont {B.}~\bibnamefont {{Farr}}},
  \bibinfo {author} {\bibfnamefont {R.}~\bibnamefont {{Hur}}}, \bibinfo
  {author} {\bibfnamefont {B.}~\bibnamefont {{Edelman}}}, \ and\ \bibinfo
  {author} {\bibfnamefont {Z.}~\bibnamefont {{Doctor}}},\ }\href@noop {}
  {\bibfield  {journal} {\bibinfo  {journal} {arXiv e-prints}\ ,\ \bibinfo
  {eid} {arXiv:2108.12044}} (\bibinfo {year} {2021})},\ \Eprint
  {http://arxiv.org/abs/2108.12044} {arXiv:2108.12044 [gr-qc]} \BibitemShut
  {NoStop}%
\bibitem [{\citenamefont {{Davis}}(2021)}]{2021CQGra..38m5014D}%
  \BibitemOpen
  \bibfield  {author} {\bibinfo {author} {\bibfnamefont {D.~e.~a.}\
  \bibnamefont {{Davis}}},\ }\href {\doibase 10.1088/1361-6382/abfd85}
  {\bibfield  {journal} {\bibinfo  {journal} {Classical and Quantum Gravity}\
  }\textbf {\bibinfo {volume} {38}},\ \bibinfo {eid} {135014} (\bibinfo {year}
  {2021})},\ \Eprint {http://arxiv.org/abs/2101.11673} {arXiv:2101.11673
  [astro-ph.IM]} \BibitemShut {NoStop}%
\bibitem [{\citenamefont {{Bianchi}}\ \emph {et~al.}(2021)\citenamefont
  {{Bianchi}}, \citenamefont {{Longo}}, \citenamefont {{Valdes}}, \citenamefont
  {{Gonz{\'a}lez}},\ and\ \citenamefont {{Plastino}}}]{2021arXiv210707565B}%
  \BibitemOpen
  \bibfield  {author} {\bibinfo {author} {\bibfnamefont {S.}~\bibnamefont
  {{Bianchi}}}, \bibinfo {author} {\bibfnamefont {A.}~\bibnamefont {{Longo}}},
  \bibinfo {author} {\bibfnamefont {G.}~\bibnamefont {{Valdes}}}, \bibinfo
  {author} {\bibfnamefont {G.}~\bibnamefont {{Gonz{\'a}lez}}}, \ and\ \bibinfo
  {author} {\bibfnamefont {W.}~\bibnamefont {{Plastino}}},\ }\href@noop {}
  {\bibfield  {journal} {\bibinfo  {journal} {arXiv e-prints}\ ,\ \bibinfo
  {eid} {arXiv:2107.07565}} (\bibinfo {year} {2021})},\ \Eprint
  {http://arxiv.org/abs/2107.07565} {arXiv:2107.07565 [astro-ph.IM]}
  \BibitemShut {NoStop}%
\bibitem [{\citenamefont {{Nguyen}}(2021)}]{2021CQGra..38n5001N}%
  \BibitemOpen
  \bibfield  {author} {\bibinfo {author} {\bibfnamefont {P.~e.~a.}\
  \bibnamefont {{Nguyen}}},\ }\href {\doibase 10.1088/1361-6382/ac011a}
  {\bibfield  {journal} {\bibinfo  {journal} {Classical and Quantum Gravity}\
  }\textbf {\bibinfo {volume} {38}},\ \bibinfo {eid} {145001} (\bibinfo {year}
  {2021})},\ \Eprint {http://arxiv.org/abs/2101.09935} {arXiv:2101.09935
  [astro-ph.IM]} \BibitemShut {NoStop}%
\bibitem [{\citenamefont {{Cannon}}(2021)}]{2021SoftX..1400680C}%
  \BibitemOpen
  \bibfield  {author} {\bibinfo {author} {\bibfnamefont {K.~e.~a.}\
  \bibnamefont {{Cannon}}},\ }\href {\doibase 10.1016/j.softx.2021.100680}
  {\bibfield  {journal} {\bibinfo  {journal} {SoftwareX}\ }\textbf {\bibinfo
  {volume} {14}},\ \bibinfo {eid} {100680} (\bibinfo {year} {2021})},\ \Eprint
  {http://arxiv.org/abs/2010.05082} {arXiv:2010.05082 [astro-ph.IM]}
  \BibitemShut {NoStop}%
\bibitem [{\citenamefont {{Stachie}}\ \emph {et~al.}(2020)\citenamefont
  {{Stachie}}, \citenamefont {{Canton}}, \citenamefont {{Burns}}, \citenamefont
  {{Christensen}}, \citenamefont {{Hamburg}}, \citenamefont {{Briggs}},
  \citenamefont {{Broida}}, \citenamefont {{Goldstein}}, \citenamefont
  {{Hayes}}, \citenamefont {{Littenberg}}, \citenamefont {{Shawhan}},
  \citenamefont {{Veitch}}, \citenamefont {{Veres}},\ and\ \citenamefont
  {{Wilson-Hodge}}}]{2020CQGra..37q5001S}%
  \BibitemOpen
  \bibfield  {author} {\bibinfo {author} {\bibfnamefont {C.}~\bibnamefont
  {{Stachie}}}, \bibinfo {author} {\bibfnamefont {T.~D.}\ \bibnamefont
  {{Canton}}}, \bibinfo {author} {\bibfnamefont {E.}~\bibnamefont {{Burns}}},
  \bibinfo {author} {\bibfnamefont {N.}~\bibnamefont {{Christensen}}}, \bibinfo
  {author} {\bibfnamefont {R.}~\bibnamefont {{Hamburg}}}, \bibinfo {author}
  {\bibfnamefont {M.}~\bibnamefont {{Briggs}}}, \bibinfo {author}
  {\bibfnamefont {J.}~\bibnamefont {{Broida}}}, \bibinfo {author}
  {\bibfnamefont {A.}~\bibnamefont {{Goldstein}}}, \bibinfo {author}
  {\bibfnamefont {F.}~\bibnamefont {{Hayes}}}, \bibinfo {author} {\bibfnamefont
  {T.}~\bibnamefont {{Littenberg}}}, \bibinfo {author} {\bibfnamefont
  {P.}~\bibnamefont {{Shawhan}}}, \bibinfo {author} {\bibfnamefont
  {J.}~\bibnamefont {{Veitch}}}, \bibinfo {author} {\bibfnamefont
  {P.}~\bibnamefont {{Veres}}}, \ and\ \bibinfo {author} {\bibfnamefont
  {C.~A.}\ \bibnamefont {{Wilson-Hodge}}},\ }\href {\doibase
  10.1088/1361-6382/aba28a} {\bibfield  {journal} {\bibinfo  {journal}
  {Classical and Quantum Gravity}\ }\textbf {\bibinfo {volume} {37}},\ \bibinfo
  {eid} {175001} (\bibinfo {year} {2020})},\ \Eprint
  {http://arxiv.org/abs/2001.01462} {arXiv:2001.01462 [gr-qc]} \BibitemShut
  {NoStop}%
\bibitem [{\citenamefont {{Davis}}\ \emph {et~al.}(2020)\citenamefont
  {{Davis}}, \citenamefont {{White}},\ and\ \citenamefont
  {{Saulson}}}]{2020CQGra..37n5001D}%
  \BibitemOpen
  \bibfield  {author} {\bibinfo {author} {\bibfnamefont {D.}~\bibnamefont
  {{Davis}}}, \bibinfo {author} {\bibfnamefont {L.~V.}\ \bibnamefont
  {{White}}}, \ and\ \bibinfo {author} {\bibfnamefont {P.~R.}\ \bibnamefont
  {{Saulson}}},\ }\href {\doibase 10.1088/1361-6382/ab91e6} {\bibfield
  {journal} {\bibinfo  {journal} {Classical and Quantum Gravity}\ }\textbf
  {\bibinfo {volume} {37}},\ \bibinfo {eid} {145001} (\bibinfo {year}
  {2020})},\ \Eprint {http://arxiv.org/abs/2002.09429} {arXiv:2002.09429
  [gr-qc]} \BibitemShut {NoStop}%
\bibitem [{\citenamefont {{Essick}}\ \emph {et~al.}(2020)\citenamefont
  {{Essick}}, \citenamefont {{Godwin}}, \citenamefont {{Hanna}}, \citenamefont
  {{Blackburn}},\ and\ \citenamefont {{Katsavounidis}}}]{2020arXiv200512761E}%
  \BibitemOpen
  \bibfield  {author} {\bibinfo {author} {\bibfnamefont {R.}~\bibnamefont
  {{Essick}}}, \bibinfo {author} {\bibfnamefont {P.}~\bibnamefont {{Godwin}}},
  \bibinfo {author} {\bibfnamefont {C.}~\bibnamefont {{Hanna}}}, \bibinfo
  {author} {\bibfnamefont {L.}~\bibnamefont {{Blackburn}}}, \ and\ \bibinfo
  {author} {\bibfnamefont {E.}~\bibnamefont {{Katsavounidis}}},\ }\href@noop {}
  {\bibfield  {journal} {\bibinfo  {journal} {arXiv e-prints}\ ,\ \bibinfo
  {eid} {arXiv:2005.12761}} (\bibinfo {year} {2020})},\ \Eprint
  {http://arxiv.org/abs/2005.12761} {arXiv:2005.12761 [astro-ph.IM]}
  \BibitemShut {NoStop}%
\bibitem [{\citenamefont {{Cuoco}}(2020)}]{2020arXiv200503745C}%
  \BibitemOpen
  \bibfield  {author} {\bibinfo {author} {\bibfnamefont {E.~e.~a.}\
  \bibnamefont {{Cuoco}}},\ }\href@noop {} {\bibfield  {journal} {\bibinfo
  {journal} {arXiv e-prints}\ ,\ \bibinfo {eid} {arXiv:2005.03745}} (\bibinfo
  {year} {2020})},\ \Eprint {http://arxiv.org/abs/2005.03745} {arXiv:2005.03745
  [astro-ph.HE]} \BibitemShut {NoStop}%
\bibitem [{\citenamefont {{Razzano}}\ and\ \citenamefont
  {{Cuoco}}(2018)}]{2018CQGra..35i5016R}%
  \BibitemOpen
  \bibfield  {author} {\bibinfo {author} {\bibfnamefont {M.}~\bibnamefont
  {{Razzano}}}\ and\ \bibinfo {author} {\bibfnamefont {E.}~\bibnamefont
  {{Cuoco}}},\ }\href {\doibase 10.1088/1361-6382/aab793} {\bibfield  {journal}
  {\bibinfo  {journal} {Classical and Quantum Gravity}\ }\textbf {\bibinfo
  {volume} {35}},\ \bibinfo {eid} {095016} (\bibinfo {year} {2018})},\ \Eprint
  {http://arxiv.org/abs/1803.09933} {arXiv:1803.09933 [gr-qc]} \BibitemShut
  {NoStop}%
\bibitem [{\citenamefont {{Mukund}}\ \emph {et~al.}(2017)\citenamefont
  {{Mukund}}, \citenamefont {{Abraham}}, \citenamefont {{Kandhasamy}},
  \citenamefont {{Mitra}},\ and\ \citenamefont
  {{Philip}}}]{2017PhRvD..95j4059M}%
  \BibitemOpen
  \bibfield  {author} {\bibinfo {author} {\bibfnamefont {N.}~\bibnamefont
  {{Mukund}}}, \bibinfo {author} {\bibfnamefont {S.}~\bibnamefont {{Abraham}}},
  \bibinfo {author} {\bibfnamefont {S.}~\bibnamefont {{Kandhasamy}}}, \bibinfo
  {author} {\bibfnamefont {S.}~\bibnamefont {{Mitra}}}, \ and\ \bibinfo
  {author} {\bibfnamefont {N.~S.}\ \bibnamefont {{Philip}}},\ }\href {\doibase
  10.1103/PhysRevD.95.104059} {\bibfield  {journal} {\bibinfo  {journal}
  {\prd}\ }\textbf {\bibinfo {volume} {95}},\ \bibinfo {eid} {104059} (\bibinfo
  {year} {2017})},\ \Eprint {http://arxiv.org/abs/1609.07259} {arXiv:1609.07259
  [astro-ph.IM]} \BibitemShut {NoStop}%
\bibitem [{\citenamefont {{Zevin}}\ \emph {et~al.}(2017)\citenamefont
  {{Zevin}}, \citenamefont {{Coughlin}}, \citenamefont {{Bahaadini}},
  \citenamefont {{Besler}}, \citenamefont {{Rohani}}, \citenamefont {{Allen}},
  \citenamefont {{Cabero}}, \citenamefont {{Crowston}}, \citenamefont
  {{Katsaggelos}}, \citenamefont {{Larson}}, \citenamefont {{Lee}},
  \citenamefont {{Lintott}}, \citenamefont {{Littenberg}}, \citenamefont
  {{Lundgren}}, \citenamefont {{{\O}sterlund}}, \citenamefont {{Smith}},
  \citenamefont {{Trouille}},\ and\ \citenamefont
  {{Kalogera}}}]{2017CQGra..34f4003Z}%
  \BibitemOpen
  \bibfield  {author} {\bibinfo {author} {\bibfnamefont {M.}~\bibnamefont
  {{Zevin}}}, \bibinfo {author} {\bibfnamefont {S.}~\bibnamefont {{Coughlin}}},
  \bibinfo {author} {\bibfnamefont {S.}~\bibnamefont {{Bahaadini}}}, \bibinfo
  {author} {\bibfnamefont {E.}~\bibnamefont {{Besler}}}, \bibinfo {author}
  {\bibfnamefont {N.}~\bibnamefont {{Rohani}}}, \bibinfo {author}
  {\bibfnamefont {S.}~\bibnamefont {{Allen}}}, \bibinfo {author} {\bibfnamefont
  {M.}~\bibnamefont {{Cabero}}}, \bibinfo {author} {\bibfnamefont
  {K.}~\bibnamefont {{Crowston}}}, \bibinfo {author} {\bibfnamefont {A.~K.}\
  \bibnamefont {{Katsaggelos}}}, \bibinfo {author} {\bibfnamefont {S.~L.}\
  \bibnamefont {{Larson}}}, \bibinfo {author} {\bibfnamefont {T.~K.}\
  \bibnamefont {{Lee}}}, \bibinfo {author} {\bibfnamefont {C.}~\bibnamefont
  {{Lintott}}}, \bibinfo {author} {\bibfnamefont {T.~B.}\ \bibnamefont
  {{Littenberg}}}, \bibinfo {author} {\bibfnamefont {A.}~\bibnamefont
  {{Lundgren}}}, \bibinfo {author} {\bibfnamefont {C.}~\bibnamefont
  {{{\O}sterlund}}}, \bibinfo {author} {\bibfnamefont {J.~R.}\ \bibnamefont
  {{Smith}}}, \bibinfo {author} {\bibfnamefont {L.}~\bibnamefont {{Trouille}}},
  \ and\ \bibinfo {author} {\bibfnamefont {V.}~\bibnamefont {{Kalogera}}},\
  }\href {\doibase 10.1088/1361-6382/aa5cea} {\bibfield  {journal} {\bibinfo
  {journal} {Classical and Quantum Gravity}\ }\textbf {\bibinfo {volume}
  {34}},\ \bibinfo {eid} {064003} (\bibinfo {year} {2017})},\ \Eprint
  {http://arxiv.org/abs/1611.04596} {arXiv:1611.04596 [gr-qc]} \BibitemShut
  {NoStop}%
\bibitem [{\citenamefont {{Valdes Sanchez}}(2017)}]{2017PhDT........25V}%
  \BibitemOpen
  \bibfield  {author} {\bibinfo {author} {\bibfnamefont {G.~A.}\ \bibnamefont
  {{Valdes Sanchez}}},\ }\emph {\bibinfo {title} {{Data Analysis Techniques for
  Ligo Detector Characterization}}},\ \href@noop {} {Ph.D. thesis},\ \bibinfo
  {school} {The University of Texas at San Antonio} (\bibinfo {year}
  {2017})\BibitemShut {NoStop}%
\bibitem [{\citenamefont {{Massinger}}(2016)}]{2016PhDT.......149M}%
  \BibitemOpen
  \bibfield  {author} {\bibinfo {author} {\bibfnamefont {T.~J.}\ \bibnamefont
  {{Massinger}}},\ }\emph {\bibinfo {title} {{Detector characterization for
  advanced LIGO}}},\ \href@noop {} {Ph.D. thesis},\ \bibinfo  {school}
  {Syracuse University} (\bibinfo {year} {2016})\BibitemShut {NoStop}%
\bibitem [{\citenamefont {{Nuttall}}\ \emph {et~al.}(2015)\citenamefont
  {{Nuttall}}, \citenamefont {{Massinger}}, \citenamefont {{Areeda}},
  \citenamefont {{Betzwieser}}, \citenamefont {{Dwyer}}, \citenamefont
  {{Effler}}, \citenamefont {{Fisher}}, \citenamefont {{Fritschel}},
  \citenamefont {{Kissel}}, \citenamefont {{Lundgren}}, \citenamefont
  {{Macleod}}, \citenamefont {{Martynov}}, \citenamefont {{McIver}},
  \citenamefont {{Mullavey}}, \citenamefont {{Sigg}}, \citenamefont {{Smith}},
  \citenamefont {{Vajente}}, \citenamefont {{Williamson}},\ and\ \citenamefont
  {{Wipf}}}]{2015CQGra..32x5005N}%
  \BibitemOpen
  \bibfield  {author} {\bibinfo {author} {\bibfnamefont {L.~K.}\ \bibnamefont
  {{Nuttall}}}, \bibinfo {author} {\bibfnamefont {T.~J.}\ \bibnamefont
  {{Massinger}}}, \bibinfo {author} {\bibfnamefont {J.}~\bibnamefont
  {{Areeda}}}, \bibinfo {author} {\bibfnamefont {J.}~\bibnamefont
  {{Betzwieser}}}, \bibinfo {author} {\bibfnamefont {S.}~\bibnamefont
  {{Dwyer}}}, \bibinfo {author} {\bibfnamefont {A.}~\bibnamefont {{Effler}}},
  \bibinfo {author} {\bibfnamefont {R.~P.}\ \bibnamefont {{Fisher}}}, \bibinfo
  {author} {\bibfnamefont {P.}~\bibnamefont {{Fritschel}}}, \bibinfo {author}
  {\bibfnamefont {J.~S.}\ \bibnamefont {{Kissel}}}, \bibinfo {author}
  {\bibfnamefont {A.~P.}\ \bibnamefont {{Lundgren}}}, \bibinfo {author}
  {\bibfnamefont {D.~M.}\ \bibnamefont {{Macleod}}}, \bibinfo {author}
  {\bibfnamefont {D.}~\bibnamefont {{Martynov}}}, \bibinfo {author}
  {\bibfnamefont {J.}~\bibnamefont {{McIver}}}, \bibinfo {author}
  {\bibfnamefont {A.}~\bibnamefont {{Mullavey}}}, \bibinfo {author}
  {\bibfnamefont {D.}~\bibnamefont {{Sigg}}}, \bibinfo {author} {\bibfnamefont
  {J.~R.}\ \bibnamefont {{Smith}}}, \bibinfo {author} {\bibfnamefont
  {G.}~\bibnamefont {{Vajente}}}, \bibinfo {author} {\bibfnamefont {A.~R.}\
  \bibnamefont {{Williamson}}}, \ and\ \bibinfo {author} {\bibfnamefont
  {C.~C.}\ \bibnamefont {{Wipf}}},\ }\href {\doibase
  10.1088/0264-9381/32/24/245005} {\bibfield  {journal} {\bibinfo  {journal}
  {Classical and Quantum Gravity}\ }\textbf {\bibinfo {volume} {32}},\ \bibinfo
  {eid} {245005} (\bibinfo {year} {2015})},\ \Eprint
  {http://arxiv.org/abs/1508.07316} {arXiv:1508.07316 [gr-qc]} \BibitemShut
  {NoStop}%
\bibitem [{\citenamefont {{Biswas}}\ \emph {et~al.}(2013)\citenamefont
  {{Biswas}}, \citenamefont {{Blackburn}}, \citenamefont {{Cao}}, \citenamefont
  {{Essick}}, \citenamefont {{Hodge}}, \citenamefont {{Katsavounidis}},
  \citenamefont {{Kim}}, \citenamefont {{Kim}}, \citenamefont {{Le Bigot}},
  \citenamefont {{Lee}}, \citenamefont {{Oh}}, \citenamefont {{Oh}},
  \citenamefont {{Son}}, \citenamefont {{Tao}}, \citenamefont {{Vaulin}},\ and\
  \citenamefont {{Wang}}}]{2013PhRvD..88f2003B}%
  \BibitemOpen
  \bibfield  {author} {\bibinfo {author} {\bibfnamefont {R.}~\bibnamefont
  {{Biswas}}}, \bibinfo {author} {\bibfnamefont {L.}~\bibnamefont
  {{Blackburn}}}, \bibinfo {author} {\bibfnamefont {J.}~\bibnamefont {{Cao}}},
  \bibinfo {author} {\bibfnamefont {R.}~\bibnamefont {{Essick}}}, \bibinfo
  {author} {\bibfnamefont {K.~A.}\ \bibnamefont {{Hodge}}}, \bibinfo {author}
  {\bibfnamefont {E.}~\bibnamefont {{Katsavounidis}}}, \bibinfo {author}
  {\bibfnamefont {K.}~\bibnamefont {{Kim}}}, \bibinfo {author} {\bibfnamefont
  {Y.-M.}\ \bibnamefont {{Kim}}}, \bibinfo {author} {\bibfnamefont {E.-O.}\
  \bibnamefont {{Le Bigot}}}, \bibinfo {author} {\bibfnamefont {C.-H.}\
  \bibnamefont {{Lee}}}, \bibinfo {author} {\bibfnamefont {J.~J.}\ \bibnamefont
  {{Oh}}}, \bibinfo {author} {\bibfnamefont {S.~H.}\ \bibnamefont {{Oh}}},
  \bibinfo {author} {\bibfnamefont {E.~J.}\ \bibnamefont {{Son}}}, \bibinfo
  {author} {\bibfnamefont {Y.}~\bibnamefont {{Tao}}}, \bibinfo {author}
  {\bibfnamefont {R.}~\bibnamefont {{Vaulin}}}, \ and\ \bibinfo {author}
  {\bibfnamefont {X.}~\bibnamefont {{Wang}}},\ }\href {\doibase
  10.1103/PhysRevD.88.062003} {\bibfield  {journal} {\bibinfo  {journal}
  {\prd}\ }\textbf {\bibinfo {volume} {88}},\ \bibinfo {eid} {062003} (\bibinfo
  {year} {2013})},\ \Eprint {http://arxiv.org/abs/1303.6984} {arXiv:1303.6984
  [astro-ph.IM]} \BibitemShut {NoStop}%
\bibitem [{\citenamefont {{MacLeod}}(2013)}]{2013PhDT.......555M}%
  \BibitemOpen
  \bibfield  {author} {\bibinfo {author} {\bibfnamefont {D.}~\bibnamefont
  {{MacLeod}}},\ }\emph {\bibinfo {title} {{Improving the sensitivity of
  searches for gravitational waves from compact binary coalescences}}},\
  \href@noop {} {Ph.D. thesis},\ \bibinfo  {school} {Cardiff University (United
  Kingdom)} (\bibinfo {year} {2013})\BibitemShut {NoStop}%
\bibitem [{\citenamefont {{Aasi}}(2012)}]{2012CQGra..29o5002A}%
  \BibitemOpen
  \bibfield  {author} {\bibinfo {author} {\bibfnamefont {J.~e.~a.}\
  \bibnamefont {{Aasi}}},\ }\href {\doibase 10.1088/0264-9381/29/15/155002}
  {\bibfield  {journal} {\bibinfo  {journal} {Classical and Quantum Gravity}\
  }\textbf {\bibinfo {volume} {29}},\ \bibinfo {eid} {155002} (\bibinfo {year}
  {2012})},\ \Eprint {http://arxiv.org/abs/1203.5613} {arXiv:1203.5613 [gr-qc]}
  \BibitemShut {NoStop}%
\bibitem [{\citenamefont {{Christensen}}\ \emph {et~al.}(2010)\citenamefont
  {{Christensen}}, \citenamefont {{LIGO Scientific Collaboration}},\ and\
  \citenamefont {{Virgo Collaboration}}}]{2010CQGra..27s4010C}%
  \BibitemOpen
  \bibfield  {author} {\bibinfo {author} {\bibfnamefont {N.}~\bibnamefont
  {{Christensen}}}, \bibinfo {author} {\bibnamefont {{LIGO Scientific
  Collaboration}}}, \ and\ \bibinfo {author} {\bibnamefont {{Virgo
  Collaboration}}},\ }\href {\doibase 10.1088/0264-9381/27/19/194010}
  {\bibfield  {journal} {\bibinfo  {journal} {Classical and Quantum Gravity}\
  }\textbf {\bibinfo {volume} {27}},\ \bibinfo {eid} {194010} (\bibinfo {year}
  {2010})}\BibitemShut {NoStop}%
\bibitem [{\citenamefont {{Isogai}}\ \emph {et~al.}(2010)\citenamefont
  {{Isogai}}, \citenamefont {{LIGO Scientific Collaboration}},\ and\
  \citenamefont {{Virgo Collaboration}}}]{2010JPhCS.243a2005I}%
  \BibitemOpen
  \bibfield  {author} {\bibinfo {author} {\bibfnamefont {T.}~\bibnamefont
  {{Isogai}}}, \bibinfo {author} {\bibnamefont {{LIGO Scientific
  Collaboration}}}, \ and\ \bibinfo {author} {\bibnamefont {{Virgo
  Collaboration}}},\ }in\ \href {\doibase 10.1088/1742-6596/243/1/012005}
  {\emph {\bibinfo {booktitle} {Journal of Physics Conference Series}}},\
  \bibinfo {series} {Journal of Physics Conference Series}, Vol.\ \bibinfo
  {volume} {243}\ (\bibinfo {year} {2010})\ p.\ \bibinfo {pages}
  {012005}\BibitemShut {NoStop}%
\bibitem [{\citenamefont {{Mukherjee}}\ \emph {et~al.}(2010)\citenamefont
  {{Mukherjee}}, \citenamefont {{Obaid}},\ and\ \citenamefont
  {{Matkarimov}}}]{2010JPhCS.243a2006M}%
  \BibitemOpen
  \bibfield  {author} {\bibinfo {author} {\bibfnamefont {S.}~\bibnamefont
  {{Mukherjee}}}, \bibinfo {author} {\bibfnamefont {R.}~\bibnamefont
  {{Obaid}}}, \ and\ \bibinfo {author} {\bibfnamefont {B.}~\bibnamefont
  {{Matkarimov}}},\ }in\ \href {\doibase 10.1088/1742-6596/243/1/012006} {\emph
  {\bibinfo {booktitle} {Journal of Physics Conference Series}}},\ \bibinfo
  {series} {Journal of Physics Conference Series}, Vol.\ \bibinfo {volume}
  {243}\ (\bibinfo {year} {2010})\ p.\ \bibinfo {pages} {012006}\BibitemShut
  {NoStop}%
\bibitem [{\citenamefont {{Blackburn}}(2008)}]{2008CQGra..25r4004B}%
  \BibitemOpen
  \bibfield  {author} {\bibinfo {author} {\bibfnamefont {L.~e.~a.}\
  \bibnamefont {{Blackburn}}},\ }\href {\doibase
  10.1088/0264-9381/25/18/184004} {\bibfield  {journal} {\bibinfo  {journal}
  {Classical and Quantum Gravity}\ }\textbf {\bibinfo {volume} {25}},\ \bibinfo
  {eid} {184004} (\bibinfo {year} {2008})},\ \Eprint
  {http://arxiv.org/abs/0804.0800} {arXiv:0804.0800 [gr-qc]} \BibitemShut
  {NoStop}%
\bibitem [{\citenamefont {{Sigg}}\ \emph {et~al.}(2002)\citenamefont {{Sigg}},
  \citenamefont {{Bork}},\ and\ \citenamefont
  {{Zweizig}}}]{2002nmgm.meet.1841S}%
  \BibitemOpen
  \bibfield  {author} {\bibinfo {author} {\bibfnamefont {D.}~\bibnamefont
  {{Sigg}}}, \bibinfo {author} {\bibfnamefont {R.}~\bibnamefont {{Bork}}}, \
  and\ \bibinfo {author} {\bibfnamefont {J.}~\bibnamefont {{Zweizig}}},\ }in\
  \href {\doibase 10.1142/9789812777386\_0401} {\emph {\bibinfo {booktitle}
  {The Ninth Marcel Grossmann Meeting}}},\ \bibinfo {editor} {edited by\
  \bibinfo {editor} {\bibfnamefont {V.~G.}\ \bibnamefont {{Gurzadyan}}},
  \bibinfo {editor} {\bibfnamefont {R.~T.}\ \bibnamefont {{Jantzen}}}, \ and\
  \bibinfo {editor} {\bibfnamefont {R.}~\bibnamefont {{Ruffini}}}}\ (\bibinfo
  {year} {2002})\ pp.\ \bibinfo {pages} {1841--1842}\BibitemShut {NoStop}%
\bibitem [{\citenamefont {Gurav}\ \emph {et~al.}(2020)\citenamefont {Gurav},
  \citenamefont {Barish}, \citenamefont {Vajente},\ and\ \citenamefont
  {Papalexakis}}]{gurav2020unsupervised}%
  \BibitemOpen
  \bibfield  {author} {\bibinfo {author} {\bibfnamefont {R.}~\bibnamefont
  {Gurav}}, \bibinfo {author} {\bibfnamefont {B.}~\bibnamefont {Barish}},
  \bibinfo {author} {\bibfnamefont {G.}~\bibnamefont {Vajente}}, \ and\
  \bibinfo {author} {\bibfnamefont {E.~E.}\ \bibnamefont {Papalexakis}},\ }in\
  \href@noop {} {\emph {\bibinfo {booktitle} {AAI 2020 Fall Symposium on
  Physics-Guided AI to Accelerate Scientific Discovery}}}\ (\bibinfo {year}
  {2020})\BibitemShut {NoStop}%
\bibitem [{\citenamefont {{Robinet}}(2015)}]{Omicron}%
  \BibitemOpen
  \bibfield  {author} {\bibinfo {author} {\bibfnamefont {F.}~\bibnamefont
  {{Robinet}}},\ }\href@noop {} {\enquote {\bibinfo {title} {{Omicron: An
  Algorithm to Detect and Characterize Transient Noise in Gravitational-Wave
  Detectors}},}\ }\bibinfo {howpublished}
  {\url{https://tds.ego-gw.it/ql/?c=10651}} (\bibinfo {year}
  {2015})\BibitemShut {NoStop}%
\bibitem [{\citenamefont {Tibshirani}(1996)}]{LASSO}%
  \BibitemOpen
  \bibfield  {author} {\bibinfo {author} {\bibfnamefont {R.}~\bibnamefont
  {Tibshirani}},\ }\href {http://www.jstor.org/stable/2346178} {\bibfield
  {journal} {\bibinfo  {journal} {Journal of the Royal Statistical Society.
  Series B (Methodological)}\ }\textbf {\bibinfo {volume} {58}},\ \bibinfo
  {pages} {267} (\bibinfo {year} {1996})}\BibitemShut {NoStop}%
\bibitem [{\citenamefont {Zou}\ and\ \citenamefont {Hastie}(2005)}]{elastic}%
  \BibitemOpen
  \bibfield  {author} {\bibinfo {author} {\bibfnamefont {H.}~\bibnamefont
  {Zou}}\ and\ \bibinfo {author} {\bibfnamefont {T.}~\bibnamefont {Hastie}},\
  }\href {http://www.jstor.org/stable/3647580} {\bibfield  {journal} {\bibinfo
  {journal} {Journal of the Royal Statistical Society. Series B (Statistical
  Methodology)}\ }\textbf {\bibinfo {volume} {67}},\ \bibinfo {pages} {301}
  (\bibinfo {year} {2005})}\BibitemShut {NoStop}%
\bibitem [{\citenamefont {Tsuruoka}\ \emph {et~al.}(2009)\citenamefont
  {Tsuruoka}, \citenamefont {Tsujii},\ and\ \citenamefont
  {Ananiadou}}]{tsuruoka2009stochastic}%
  \BibitemOpen
  \bibfield  {author} {\bibinfo {author} {\bibfnamefont {Y.}~\bibnamefont
  {Tsuruoka}}, \bibinfo {author} {\bibfnamefont {J.}~\bibnamefont {Tsujii}}, \
  and\ \bibinfo {author} {\bibfnamefont {S.}~\bibnamefont {Ananiadou}},\ }in\
  \href@noop {} {\emph {\bibinfo {booktitle} {Proceedings of the Joint
  Conference of the 47th Annual Meeting of the ACL and the 4th International
  Joint Conference on Natural Language Processing of the AFNLP}}}\ (\bibinfo
  {year} {2009})\ pp.\ \bibinfo {pages} {477--485}\BibitemShut {NoStop}%
\bibitem [{\citenamefont {Ioffe}\ and\ \citenamefont
  {Szegedy}(2015)}]{BatchNorm}%
  \BibitemOpen
  \bibfield  {author} {\bibinfo {author} {\bibfnamefont {S.}~\bibnamefont
  {Ioffe}}\ and\ \bibinfo {author} {\bibfnamefont {C.}~\bibnamefont
  {Szegedy}},\ }in\ \href@noop {} {\emph {\bibinfo {booktitle} {International
  conference on machine learning}}}\ (\bibinfo {organization} {PMLR},\ \bibinfo
  {year} {2015})\ pp.\ \bibinfo {pages} {448--456}\BibitemShut {NoStop}%
\bibitem [{\citenamefont {Yan}\ \emph {et~al.}(2022)\citenamefont {Yan},
  \citenamefont {Avagyan}, \citenamefont {Colgan}, \citenamefont {Veske},
  \citenamefont {Bartos}, \citenamefont {Wright}, \citenamefont {M\'arka},\
  and\ \citenamefont {M\'arka}}]{YCMMW-new}%
  \BibitemOpen
  \bibfield  {author} {\bibinfo {author} {\bibfnamefont {J.}~\bibnamefont
  {Yan}}, \bibinfo {author} {\bibfnamefont {M.}~\bibnamefont {Avagyan}},
  \bibinfo {author} {\bibfnamefont {R.~E.}\ \bibnamefont {Colgan}}, \bibinfo
  {author} {\bibfnamefont {D.~b.~u.}\ \bibnamefont {Veske}}, \bibinfo {author}
  {\bibfnamefont {I.}~\bibnamefont {Bartos}}, \bibinfo {author} {\bibfnamefont
  {J.}~\bibnamefont {Wright}}, \bibinfo {author} {\bibfnamefont
  {Z.}~\bibnamefont {M\'arka}}, \ and\ \bibinfo {author} {\bibfnamefont
  {S.}~\bibnamefont {M\'arka}},\ }\href {\doibase 10.1103/PhysRevD.105.043006}
  {\bibfield  {journal} {\bibinfo  {journal} {Phys. Rev. D}\ }\textbf {\bibinfo
  {volume} {105}},\ \bibinfo {pages} {043006} (\bibinfo {year}
  {2022})}\BibitemShut {NoStop}%
\bibitem [{\citenamefont {He}\ \emph {et~al.}(2015)\citenamefont {He},
  \citenamefont {Zhang}, \citenamefont {Ren},\ and\ \citenamefont
  {Sun}}]{he2015delving}%
  \BibitemOpen
  \bibfield  {author} {\bibinfo {author} {\bibfnamefont {K.}~\bibnamefont
  {He}}, \bibinfo {author} {\bibfnamefont {X.}~\bibnamefont {Zhang}}, \bibinfo
  {author} {\bibfnamefont {S.}~\bibnamefont {Ren}}, \ and\ \bibinfo {author}
  {\bibfnamefont {J.}~\bibnamefont {Sun}},\ }in\ \href@noop {} {\emph {\bibinfo
  {booktitle} {Proceedings of the IEEE international conference on computer
  vision}}}\ (\bibinfo {year} {2015})\ pp.\ \bibinfo {pages}
  {1026--1034}\BibitemShut {NoStop}%
\bibitem [{\citenamefont {Kingma}\ and\ \citenamefont {Ba}(2014)}]{adam}%
  \BibitemOpen
  \bibfield  {author} {\bibinfo {author} {\bibfnamefont {D.~P.}\ \bibnamefont
  {Kingma}}\ and\ \bibinfo {author} {\bibfnamefont {J.}~\bibnamefont {Ba}},\
  }\href@noop {} {\bibfield  {journal} {\bibinfo  {journal} {arXiv preprint
  arXiv:1412.6980}\ } (\bibinfo {year} {2014})}\BibitemShut {NoStop}%
\bibitem [{\citenamefont {Paszke}\ \emph {et~al.}(2019)\citenamefont {Paszke},
  \citenamefont {Gross}, \citenamefont {Massa}, \citenamefont {Lerer},
  \citenamefont {Bradbury}, \citenamefont {Chanan}, \citenamefont {Killeen},
  \citenamefont {Lin}, \citenamefont {Gimelshein}, \citenamefont {Antiga},
  \citenamefont {Desmaison}, \citenamefont {Kopf}, \citenamefont {Yang},
  \citenamefont {DeVito}, \citenamefont {Raison}, \citenamefont {Tejani},
  \citenamefont {Chilamkurthy}, \citenamefont {Steiner}, \citenamefont {Fang},
  \citenamefont {Bai},\ and\ \citenamefont {Chintala}}]{pytorch}%
  \BibitemOpen
  \bibfield  {author} {\bibinfo {author} {\bibfnamefont {A.}~\bibnamefont
  {Paszke}}, \bibinfo {author} {\bibfnamefont {S.}~\bibnamefont {Gross}},
  \bibinfo {author} {\bibfnamefont {F.}~\bibnamefont {Massa}}, \bibinfo
  {author} {\bibfnamefont {A.}~\bibnamefont {Lerer}}, \bibinfo {author}
  {\bibfnamefont {J.}~\bibnamefont {Bradbury}}, \bibinfo {author}
  {\bibfnamefont {G.}~\bibnamefont {Chanan}}, \bibinfo {author} {\bibfnamefont
  {T.}~\bibnamefont {Killeen}}, \bibinfo {author} {\bibfnamefont
  {Z.}~\bibnamefont {Lin}}, \bibinfo {author} {\bibfnamefont {N.}~\bibnamefont
  {Gimelshein}}, \bibinfo {author} {\bibfnamefont {L.}~\bibnamefont {Antiga}},
  \bibinfo {author} {\bibfnamefont {A.}~\bibnamefont {Desmaison}}, \bibinfo
  {author} {\bibfnamefont {A.}~\bibnamefont {Kopf}}, \bibinfo {author}
  {\bibfnamefont {E.}~\bibnamefont {Yang}}, \bibinfo {author} {\bibfnamefont
  {Z.}~\bibnamefont {DeVito}}, \bibinfo {author} {\bibfnamefont
  {M.}~\bibnamefont {Raison}}, \bibinfo {author} {\bibfnamefont
  {A.}~\bibnamefont {Tejani}}, \bibinfo {author} {\bibfnamefont
  {S.}~\bibnamefont {Chilamkurthy}}, \bibinfo {author} {\bibfnamefont
  {B.}~\bibnamefont {Steiner}}, \bibinfo {author} {\bibfnamefont
  {L.}~\bibnamefont {Fang}}, \bibinfo {author} {\bibfnamefont {J.}~\bibnamefont
  {Bai}}, \ and\ \bibinfo {author} {\bibfnamefont {S.}~\bibnamefont
  {Chintala}},\ }in\ \href
  {http://papers.neurips.cc/paper/9015-pytorch-an-imperative-style-high-performance-deep-learning-library.pdf}
  {\emph {\bibinfo {booktitle} {Advances in Neural Information Processing
  Systems 32}}},\ \bibinfo {editor} {edited by\ \bibinfo {editor}
  {\bibfnamefont {H.}~\bibnamefont {Wallach}}, \bibinfo {editor} {\bibfnamefont
  {H.}~\bibnamefont {Larochelle}}, \bibinfo {editor} {\bibfnamefont
  {A.}~\bibnamefont {Beygelzimer}}, \bibinfo {editor} {\bibfnamefont
  {F.}~\bibnamefont {d\textquotesingle Alch\'{e}-Buc}}, \bibinfo {editor}
  {\bibfnamefont {E.}~\bibnamefont {Fox}}, \ and\ \bibinfo {editor}
  {\bibfnamefont {R.}~\bibnamefont {Garnett}}}\ (\bibinfo  {publisher} {Curran
  Associates, Inc.},\ \bibinfo {year} {2019})\ pp.\ \bibinfo {pages}
  {8024--8035}\BibitemShut {NoStop}%
\bibitem [{\citenamefont {Hoff}(2017)}]{hoff2017lasso}%
  \BibitemOpen
  \bibfield  {author} {\bibinfo {author} {\bibfnamefont {P.~D.}\ \bibnamefont
  {Hoff}},\ }\href@noop {} {\enquote {\bibinfo {title} {Lasso, fractional norm
  and structured sparse estimation using a hadamard product parametrization},}\
  } (\bibinfo {year} {2017}),\ \Eprint {http://arxiv.org/abs/1611.00040}
  {arXiv:1611.00040 [stat.CO]} \BibitemShut {NoStop}%
\bibitem [{\citenamefont {Zhao}\ \emph {et~al.}(2019)\citenamefont {Zhao},
  \citenamefont {Yang},\ and\ \citenamefont {He}}]{zhao2019implicit}%
  \BibitemOpen
  \bibfield  {author} {\bibinfo {author} {\bibfnamefont {P.}~\bibnamefont
  {Zhao}}, \bibinfo {author} {\bibfnamefont {Y.}~\bibnamefont {Yang}}, \ and\
  \bibinfo {author} {\bibfnamefont {Q.-C.}\ \bibnamefont {He}},\ }\href@noop {}
  {\enquote {\bibinfo {title} {Implicit regularization via hadamard product
  over-parametrization in high-dimensional linear regression},}\ } (\bibinfo
  {year} {2019}),\ \Eprint {http://arxiv.org/abs/1903.09367} {arXiv:1903.09367
  [math.ST]} \BibitemShut {NoStop}%
\end{thebibliography}%
